\documentclass[sigconf]{acmart} 
\usepackage{arydshln}

\AtBeginDocument{%
  \providecommand\BibTeX{{%
    \normalfont B\kern-0.5em{\scshape i\kern-0.25em b}\kern-0.8em\TeX}}}

\setcopyright{acmlicensed}
\copyrightyear{2024}
\acmYear{2024}
\acmDOI{XXXXXXX.XXXXXXX}

\acmISBN{978-1-4503-XXXX-X/18/06}

\begin{document}

\title{AnyLoss: Transforming Classification Metrics into Loss Functions}

\author{Doheon Han}
\affiliation{%
  \institution{University of Notre Dame}
  \streetaddress{Holy Cross Drive}
  \city{Notre Dame}
  \state{Indiana}
  \country{USA}
  \postcode{46556}}
\email{dhan6@nd.edu}

\author{Nuno Moniz}
\affiliation{%
  \institution{University of Notre Dame}
  \streetaddress{Holy Cross Drive}
  \city{Notre Dame}
  \state{Indiana}
  \country{USA}
  \postcode{46556}}
\email{nmoniz2@nd.edu}

\author{Nitesh V Chawla}
\affiliation{%
  \institution{University of Notre Dame}
  \streetaddress{Holy Cross Drive}
  \city{Notre Dame}
  \state{Indiana}
  \country{USA}
  \postcode{46556}}
\email{nchawla@nd.edu}

\renewcommand{\shortauthors}{Doheon, et al.}

\begin{abstract}
Many evaluation metrics can be used to assess the performance of models in binary classification tasks.
However, most of them are derived from a confusion matrix in a non-differentiable form, making it very difficult to generate a differentiable loss function that could directly optimize them. 
The lack of solutions to bridge this challenge not only hinders our ability to solve difficult tasks, such as imbalanced learning, but also requires the deployment of computationally expensive hyperparameter search processes in model selection.
In this paper, we propose a general-purpose approach that transforms any confusion matrix-based metric into a loss function,  \textit{AnyLoss}, that is available in optimization processes. 
To this end, we use an approximation function to make a confusion matrix represented in a differentiable form, and this approach enables any confusion matrix-based metric to be directly used as a loss function. 
The mechanism of the approximation function is provided to ensure its operability and the differentiability of our loss functions is proved by suggesting their derivatives.
We conduct extensive experiments under diverse neural networks with many datasets, and we demonstrate their general availability to target any confusion matrix-based metrics.
Our method, especially, shows outstanding achievements in dealing with imbalanced datasets, and its competitive learning speed, compared to multiple baseline models, underscores its efficiency.
\end{abstract}





\maketitle
\section{Introduction}
Model evaluation is a vital aspect of machine learning, and selecting an appropriate evaluation metric is a challenging step due to the availability of multiple metrics~\cite{raschka2018model}. 
This selection process can be considered a goal in model development since it is closely related to the task objectives defined by users and the characteristics of the data domains. 
Most of them, such as accuracy or F$_{\beta}$ scores, are derived from a confusion matrix~\cite{lever2016classification}, obtained from operations over discrete values, i.e., labels.
Such labels are generated from a threshold function that transforms continuous values, i.e., class probabilities, into discrete values, meaning that the confusion matrix is in a non-differentiable form.
Importantly, this implies that confusion matrix-based metrics cannot be used as a goal, i.e., a loss function, in model development, even though the ultimate goal of the learning is to achieve a better score for the selected metric.

\begin{figure}[!t]
  \centering
  \includegraphics[width=\linewidth]{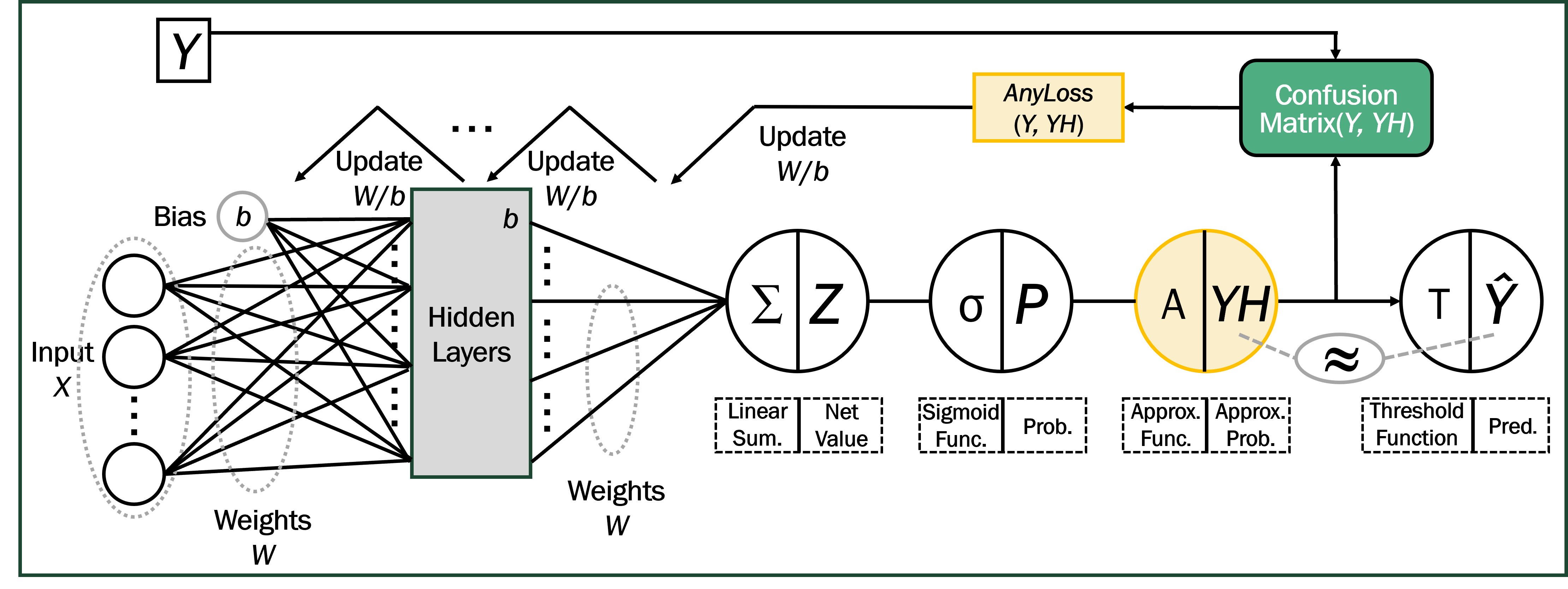}
  \caption{Our method in the multi-layer perceptron (MLP) structure. Input \textbf{X} and weights \textbf{W} generate the net value \textbf{Z}, and the sigmoid function \textit{$\sigma$} transform the net value into the class probability \textbf{P}. The approximation function \textit{$A$} generates \textbf{YH} by amplifying the probability \textbf{P}. The confusion matrix is constructed in a differentiable form by the ground truth \textbf{Y} and the \textbf{YH}. Consequently, our loss function, \textit{AnyLoss}, aimed at any confusion matrix-based metric, is available in a differentiable form.}
  \label{fig:method_mlp}
\end{figure}

Mean Squared Error (MSE), L2 loss~\cite{chai2014root}, and Binary Cross Entropy (BCE) are generally used as loss functions in neural networks~\cite{janocha2017loss}. However, they cannot directly target the desired metric, although indirectly aiming at a higher accuracy, i.e., a lower error rate.
Therefore, several strategies have been suggested to meet the unique needs of classification tasks.
The thresholding strategy finds the optimal threshold value to get a desired score of a chosen evaluation metric~\cite{lipton2014optimal, zou2016finding}, the data pre-processing strategy deals with raw data issues such as inconsistencies, imbalance, etc.~\cite{benhar2020data}, and the surrogate loss function strategy provides a new loss function that aimed at the evaluation metric scores~\cite{lee2021surrogate, busa2015online, benedict2021sigmoidf1, Huang_2021_CVPR, MARCHETTI2022108913}. 
However, the thresholding strategies have some issues like the precision-recall trade-off~\cite{zou2016finding, yang2001study}, and the data pre-processing, such as data resampling, has some issues like overfitting or data losing ~\cite{ganganwar2012overview, yuan2023review}.
The surrogate loss function strategy that is comparatively free from the aforementioned issues has been studied, and they are discussed more in the related work section.

In this paper, we introduce a general-purpose method to generate a surrogate loss function, \textit{AnyLoss}, directly aimed at any desired confusion matrix-based evaluation metric.
Our method contains an approximation function that amplifies the class probability to build a confusion matrix in a differentiable form.
This allows the estimated metric scores to be calculated before updating weights.
In a neural network for binary classification, class probability is generated from the sigmoid function after the output layer, i.e., the probability of being classified as a positive case.
The approximation function amplifies the probability to close to $0$ or $1$ to be regarded as the predicted label. The approximated probability can be used with the ground truth label to build a confusion matrix, and the evaluation metrics calculated from the matrix are represented in a differentiable form.
The metric scores from the confusion matrix are similar to those of an actual confusion matrix when the approximated probability is close enough to the predicted label.
Consequently, our method can use any confusion matrix-based evaluation metric as a loss function to optimize learning processes.

To show the operability of our method in neural networks, we provide mathematical and experimental analysis for the approximation function and prove the differentiability of loss functions by suggesting their derivatives. 
Extensive experiments using several datasets are conducted to assess the availability of our method in single- and multi-layer perceptron structures.
The performance of the experiments shows our method's ability for generalization, indicating that our method can indeed be used with any confusion matrix-based evaluation metric.
Our method, especially, shows considerable performance with imbalanced datasets since our method is available for any desired metric.
In addition, our method's learning speed is competitive with the baseline models, as demonstrated by experimental results.

Section 2 introduces related works, and Section 3 explains our method in detail, including the form of the approximation function and the derivatives of the selected loss functions. In Section 4, we demonstrate the effectiveness of our method through experimental results using various datasets. Section 5 summarizes the strengths and limitations of our method and concludes this paper.

\section{Related Works}
There have been approaches to optimize the evaluation metric scores in classification by adopting surrogate loss functions.
Some introduce a method available for a certain metric~\cite{lee2021surrogate, busa2015online, benedict2021sigmoidf1}; in this case, a unique surrogate loss is suggested for the targeted evaluation metric.
Others provide methods that can generally be used for any evaluation metric~\cite{Huang_2021_CVPR, MARCHETTI2022108913}, and ours falls into this category.
Regardless of strategies, methods such as approximating the gradient of the score, assuming the score in a stochastic setting, or constructing a temporary confusion matrix have mostly been considered.

\subsection{Specific Metric}
Most works targeting a specific metric have been pursued optimizing the F$_{\beta}$ score.
Lee at el. ~\cite{lee2021surrogate} propose a surrogate loss for the F$_{\beta}$ score inspired by the BCE gradient conditions and the mean absolute error loss.
It is derived from the derivative condition of the F$_{\beta}$ score by approximating its gradient path during learning, and its form is similar to the BCE loss.
Busa-Fekete et al. ~\cite{busa2015online} introduce a method to optimize F$_{1}$ score in an online learning environment, and the method can maximize the score on the population level with the partially observed data when the i.i.d. setting is assumed.
They consider the problem in a stochastic setting and represent the F$_{1}$ score as a function of a variable threshold.
Benedict et al. ~\cite{benedict2021sigmoidf1} introduce a surrogate loss for the F$_{1}$ score in multi-label classification.
They applied the method to their study's text and image datasets, but it is also available for general classification tasks.
They suggest a surrogate loss function for the F$_{1}$ score using the smoothed probability obtained from the sigmoid function.
The works are uniquely designed for each desired metric, so the approach's availability is limited in the area targeting a specific metric.
The next shows more general work available for most evaluation metrics.

\subsection{General Metrics}
Huang et al.~\cite{Huang_2021_CVPR} suggest the 'MetricOpt' method, which learns the approximate metric gradient for a given surrogate loss function.
This method was devised for computer vision tasks, such as image classification, but also applies to non-image data. 
It shows improved results in diverse metrics such as the miss-classification rate, Jaccard index, and F$_{1}$ score but slower than standard loss optimization since it includes finetuning steps.
This method is generally more intricate because it operates through additional processes, such as meta-learning a value function and pre-training a main model.
Marchetti et al. ~\cite{MARCHETTI2022108913} present `Score-Oriented Loss' functions for binary classification tasks. 
They try probabilistic methodology, which sets the threshold used for prediction to a continuous random variable.
Based on the assumption, a probabilistic confusion matrix is constructed, and any desired metric can be derived from it.
This method is similar to ours in constructing a confusion matrix to generate any desired evaluation metric. However, it has additional processes, such as selecting parameters: the probability distribution, the uniform or cosine-raised distribution, and two parameters $\mu$ and $\delta$ have to be chosen for the threshold. 
This method performs well with their optimized settings compared to the Cross-Entropy and the Kullback-Leibler divergence.
This method has a more complicated process than ours, which can increase learning time. The appropriate distribution and corresponding parameters must be found for each task since performance depends on them.
Compared to our simpler approach, this method has a slower learning speed and includes a relatively time-consuming process of deciding on a setting for every task, corroborated by experimental results.

\section{Method}
Our method in a multi-layer perception (MLP) structure is shown in Fig.~\ref{fig:method_mlp}.
Input data \textbf{X} = \{$\textbf{x}_{i}$|$\textbf{x}_{i}$ = $\{ x_{ij} | x_{ij} \in \mathbb{R}, j=1,\cdots,m $\}, $i=1,\cdots,n$\} with the number of samples \textit{n} and the number of features \textit{m}, and the weights \textbf{W} = \{$w_{j}$|$w_{j} \in \mathbb{R}$, $j=0,\cdots,m$\} generate the net value \textbf{Z} = \{$z_{i}$|$z_{i} = w_{0} + \sum_{j=1}^{m} x_{ij} \cdot w_{j}$, $i=1,\cdots,n$\}.
The class probability \textbf{P} = \{$p_{i}$|$p_{i} = \sigma(z_{i})$, $i=1,\cdots,n$\} is generated from the sigmoid function $\sigma$.
The approximated probability \textbf{YH} = \{$yh_{i}$|$yh_{i} = A(p_{i})$, $i=1,\cdots,n$\} is calculated from the approximation function \textit{A}.
Each value of \textbf{YH} is very close to 0 or 1 so that it can be considered as the predicted labels $\widehat{\textbf{Y}}$ = \{$\widehat{y}_{i}$|$\widehat{y}_{i} =  T(p_{i})$, $i=1,\cdots,n$\}.
The \textbf{YH} and the ground truths \textbf{Y} = \{$y_{i}$|$y_{i} \in \{0,1\}$, $i=1,\cdots,n$\} can build a confusion matrix, and any evaluation metrics can be generated.
When a desired evaluation metric is set as a loss function, \textit{AnyLoss}, the weights will be updated to optimize the loss function.

\subsection{Approximation}
The role of the approximation function is, in brief, \textbf{``To make the \textit{A(p$_{i}$)} close to 0 or 1 but not converged to exact 0 or 1 with the given \textit{p$_{i}$}"}.
The mathematical form of the approximation function is shown as (\ref{eq:approximation}).
The amplifying scale \textit{L} is a positive real number, and \textit{$p_{i}$} is the given class probability after the sigmoid function.
For its operation, the function \textit{A($p_{i}$)} has to meet the two conditions, which decide the amplifying scale \textit{L}.
The two conditions are given in mathematical forms with examples, followed by the analysis for the proper range of the amplifying scale \textit{L}.
\begin{equation}
  A(p_{i}) = \frac{1} {1 + e^{-L(p_{i}-0.5)}}
  \label{eq:approximation}
\end{equation}

\begin{figure}[ht]
  \centering
  \includegraphics[width=.23\textwidth]{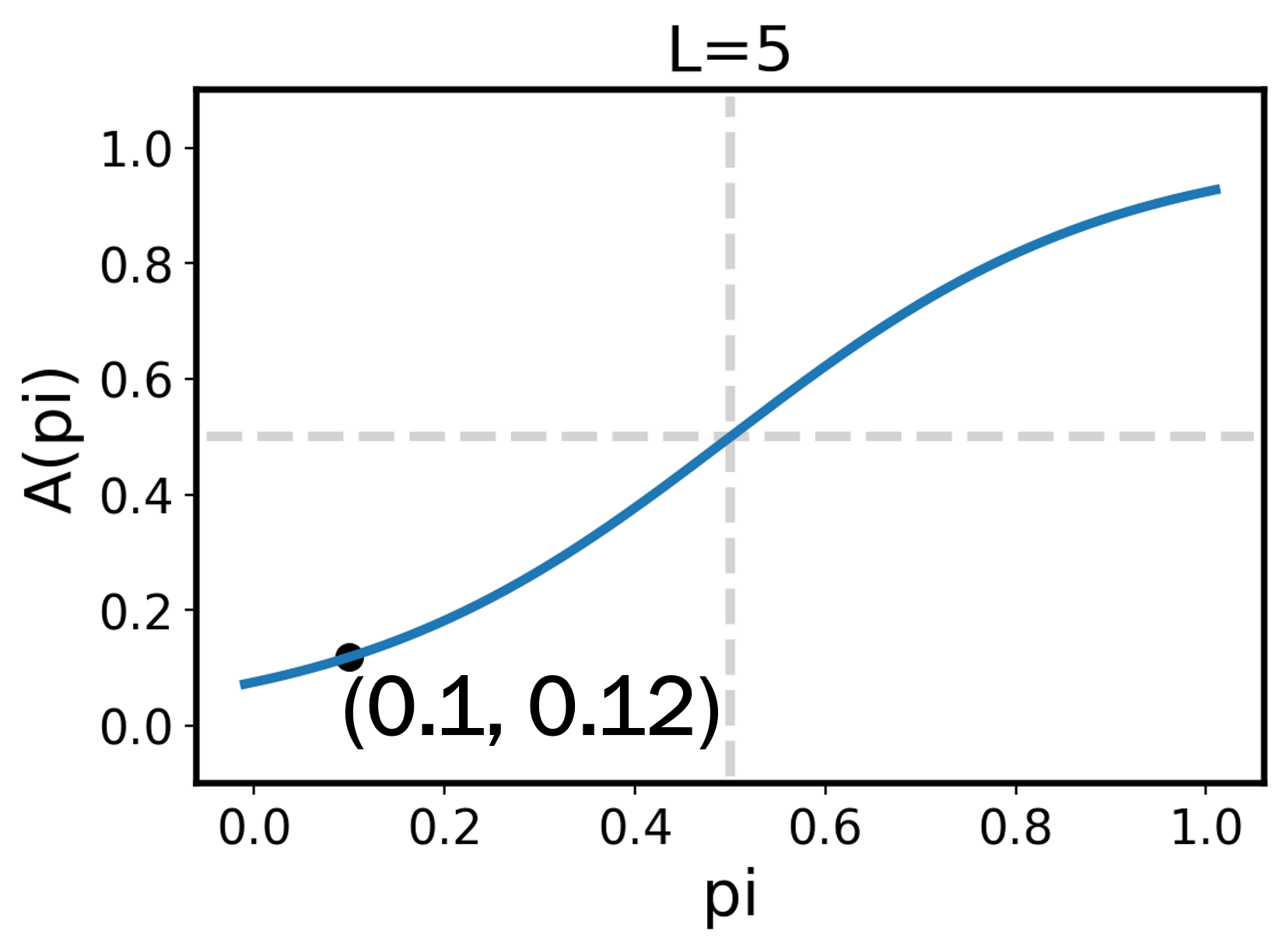}\hfill
  \includegraphics[width=.23\textwidth]{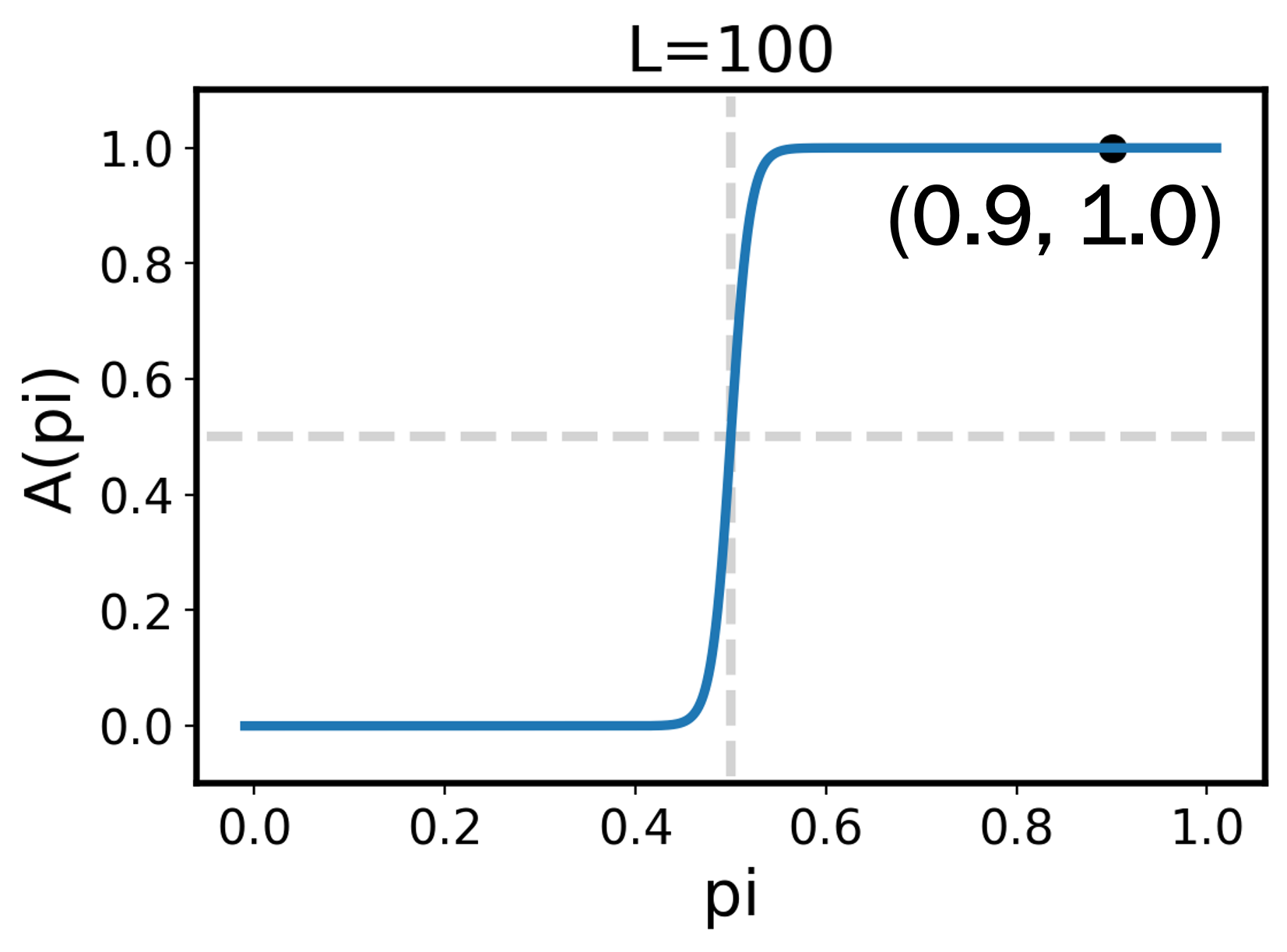}
  \caption{The approximation function \textit{A($p_{i}$)} with 2 different \textit{L} values. On the left, a smaller \textit{L}, the \textit{A($p_{i}$=0.1)} should be smaller than 0.1, but it generates 0.12. On the right, a larger \textit{L}, the \textit{A($p_{i}$=0.9)} converges to 1.0.}
\label{fig:approximation_L}
\end{figure}

\textbf{1$^{st}$ Condition: Amplifier}. 
The approximation function should be able to make the \textit{A(p$_{i}$)} closer to 1, indicating a positive label, if the given \textit{$p_{i}$} is closer to 1 than 0, and vice versa.
This process amplifies the class probability \textit{$p_{i}$} so that the approximated probability \textit{A(p$_{i}$)} can be regarded as actual labels. 
The mathematical definition of the first condition is as (\ref{eq:1stcond}).
\begin{equation}
  |A(p_{i})-0.5| \geq |p_{i}-0.5| 
  \quad OR \quad
    \left\{\begin{matrix}
        A(p_{i}) \geq p_{i}, & p_{i} \geq 0.5\\
        A(p_{i}) \leq p_{i}, & p_{i} \leq 0.5
    \end{matrix}\right.
  \label{eq:1stcond}
\end{equation}
If the class probability \textit{$p_{i}$} is larger than 0.5, the \textit{A(p$_{i}$)} should be larger than the \textit{$p_{i}$} and close to 1.
For instance, if \textit{$p_{i}$} is 0.7, the approximation function should generate the \textit{A(p$_{i}$)} larger than 0.7 and close to 1 so that we can assume it as an actual label `1'. 
Reversely, if \textit{$p_{i}$} is 0.2, the approximation function should generate the \textit{A(p$_{i}$)} smaller than 0.2 and close to 0 so that we can assume it as an actual label `0'.
However, the approximation function with a small value of \textit{L} cannot amplify the \textit{$p_{i}$} in some areas.
Fig.~\ref{fig:approximation_L} shows how a small \textit{L} violates the condition.
The X-axis and Y-axis represent each \textit{$p_{i}$} and corresponding \textit{A($p_{i}$)}.
On the left graph, the \textit{L} is five, and the slope is too gentle, so the function does not work as an amplifier in some areas.
For example, the \textit{A(p$_{i}$)} is rather larger than the corresponding \textit{$p_{i}$} on the point (0.1, 0.12) although \textit{$p_{i}$} is smaller than 0.5.
The approximation function should generate a value smaller than 0.1, but it does not.
A larger \textit{L} makes the function shape steeper. Therefore, this issue will be solved with a larger \textit{L}.
This shows that there exists a minimum value of \textit{L} to ensure the operation of the approximation function.

\textbf{2$^{nd}$ Condition: No 0/1}. 
Each value of \textit{A($p_{i}$)} should be between $0$ and $1$ as described in (\ref{eq:2ndcond}).
\begin{equation}
  0 < A(p_{i}) < 1
  \label{eq:2ndcond}
\end{equation}
If \textit{A($p_{i}$)} values become $0$ or $1$, of course, it would guarantee a more accurate calculation of evaluation metric scores since they are the same as the actual labels.
However, it can cause a serious problem, `No More Update', because of the form of the partial derivative of \textit{A($p_{i}$)} as shown in (\ref{eq:derivative_A}).
The function \textit{A($p_{i}$)} is part of a loss function because the loss function is made with the confusion matrix entries, and it includes \textbf{\textit{YH}} which is produced by the function \textit{A($p_{i}$)}. 
So the partial derivative of the loss function includes the partial derivative of \textit{A($p_{i}$)}.
The derivative contains both \textit{A($p_{i}$)} and \textit{(1-A($p_{i}$))} terms, so any case of \textit{A($p_{i}$)} = 0 or 1 will make the partial derivative 0.
The right graph in Fig.~\ref{fig:approximation_L} shows that the value of \textit{A($p_{i}$)} can be converged to $0$ or $1$.
The \textit{L} is $100$, and the slope is steep. 
The point (0.9, 1.0) shows the \textit{A($p_{i}$=0.9)} value has converged to 1.
Mathematically, the approximation function is not converged to exact 0 or 1, but it does in a machine~\cite{goldberg1991every}.
Therefore, a smaller \textit{L} should be considered in this case, which implies that a maximum value of \textit{L} should exist to ensure the operation of the approximation function.

\begin{equation}
  \frac{\partial A(p_{i})} {\partial p_{i}} = L\cdot A(p_{i})\cdot(1-A(p_{i}))
  \label{eq:derivative_A}
\end{equation}

\begin{figure}[ht]
  \centering
  \includegraphics[width=.23\textwidth]{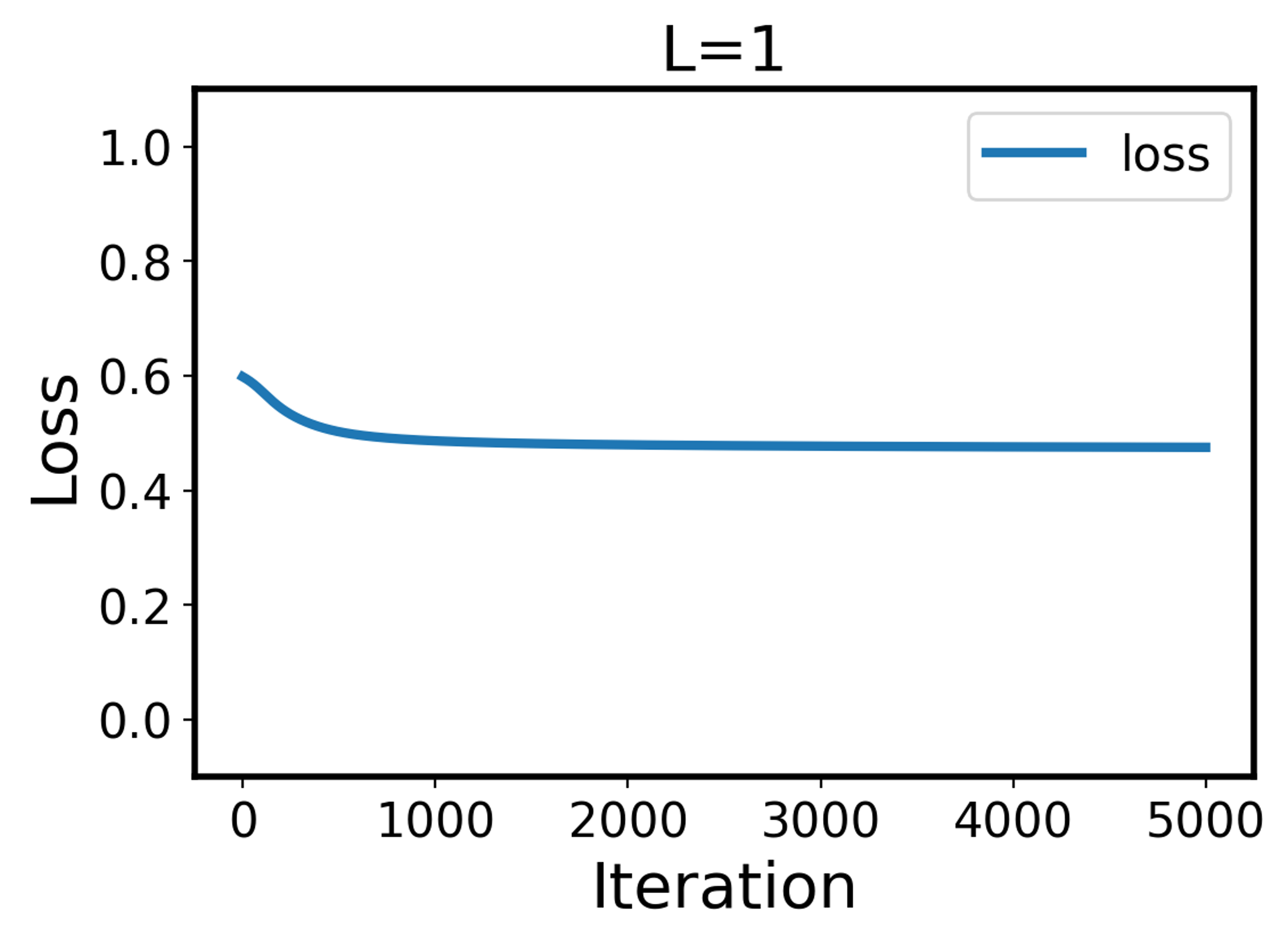}\hfill
  \includegraphics[width=.23\textwidth]{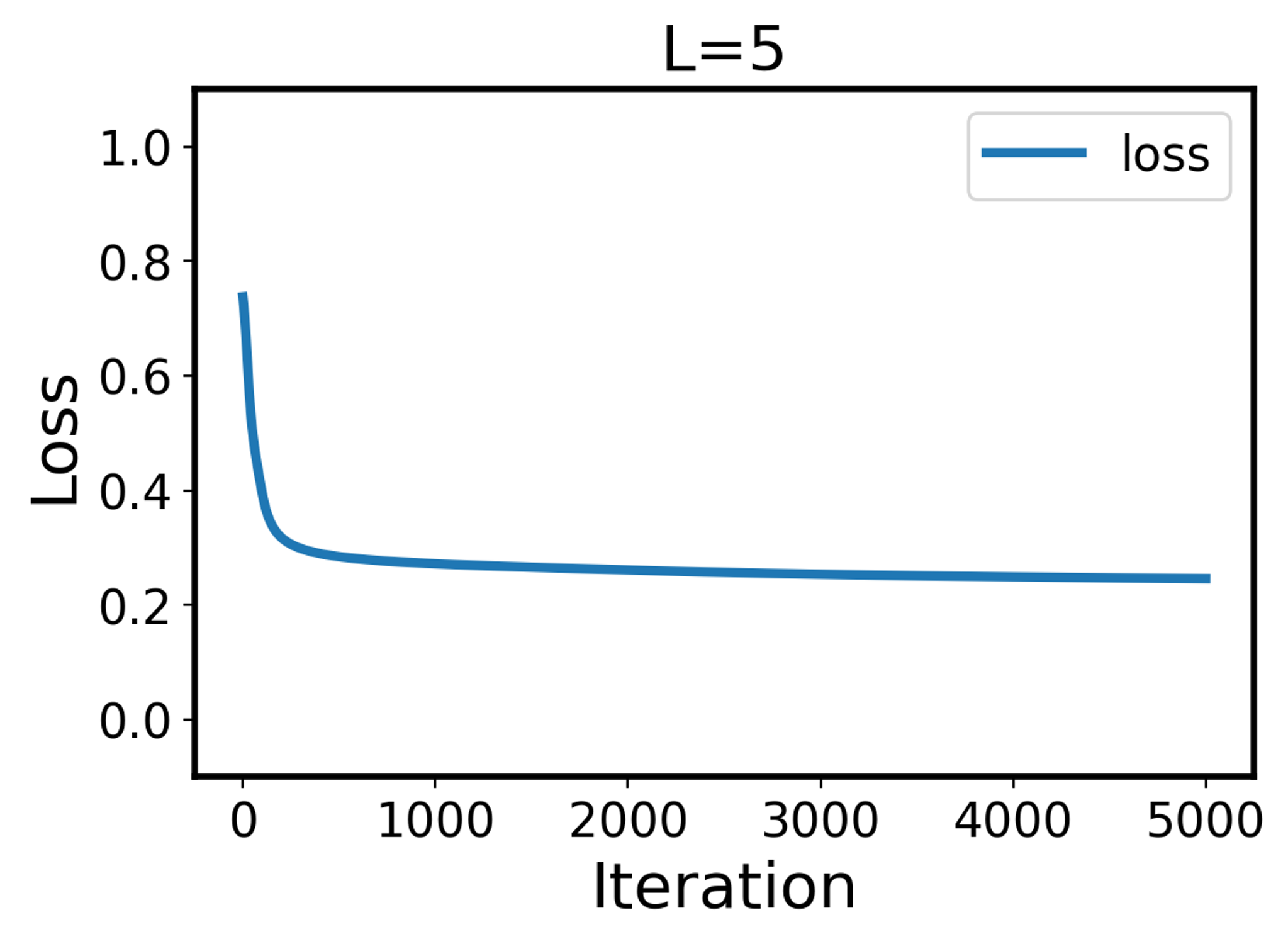}
  \includegraphics[width=.23\textwidth]{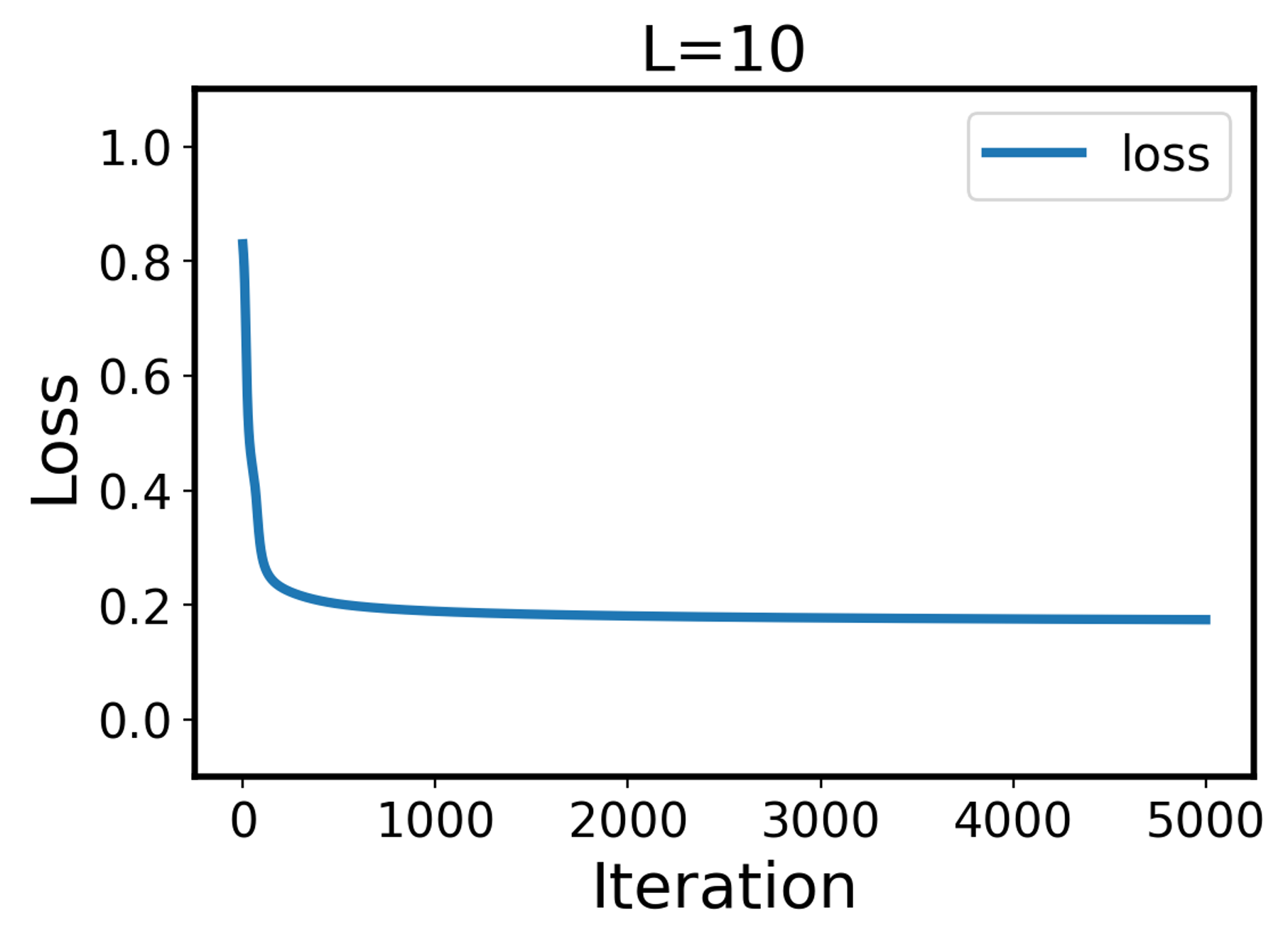}\hfill
  \includegraphics[width=.23\textwidth]{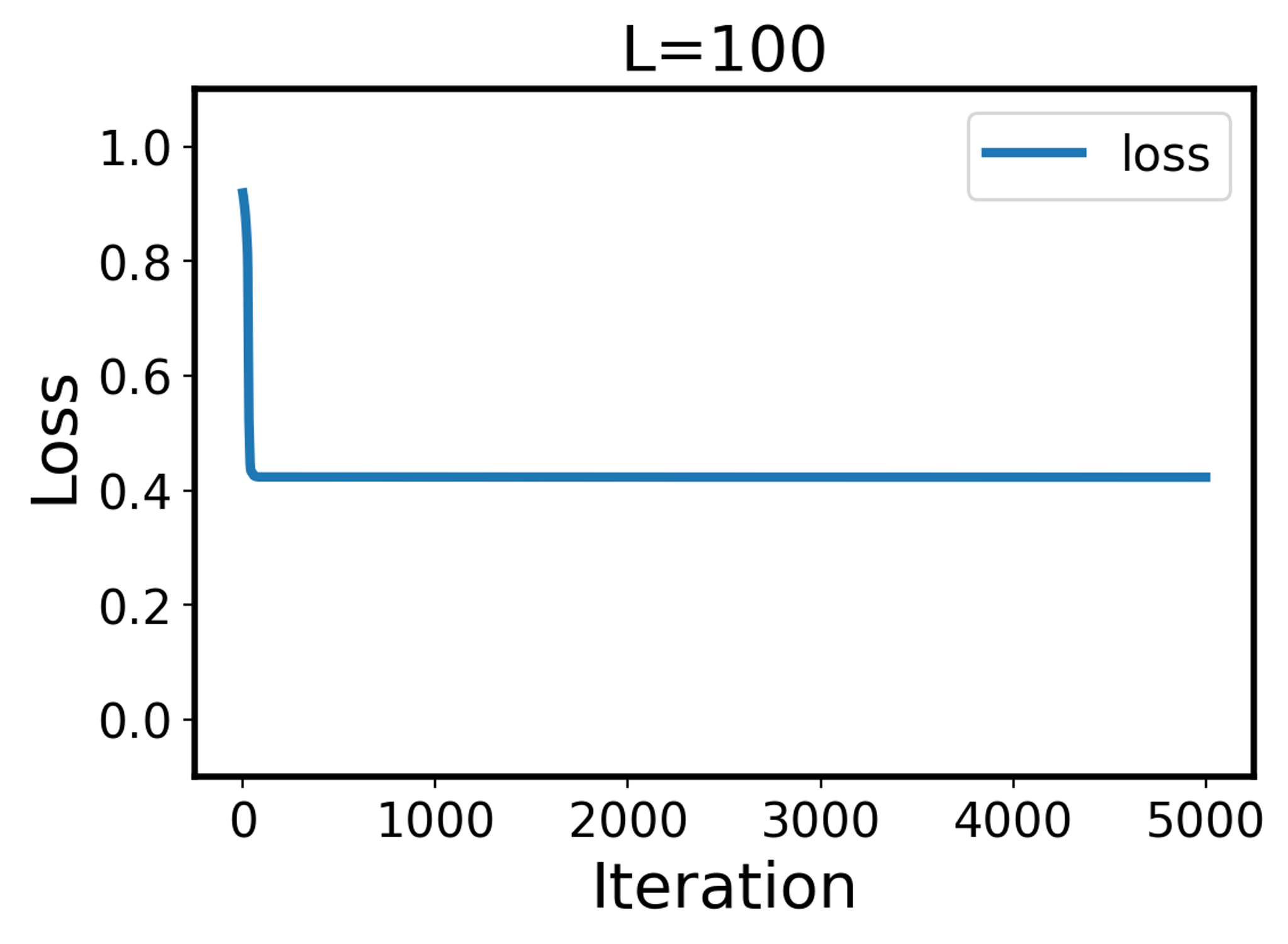}
  \caption{Learning curves with 4 different \textit{L} values. A model is not learning with a small \textit{L} but is better with a larger \textit{L}. However, a model stops learning if a \textit{L} is too large.}
\label{fig:approximation_newL}
\end{figure}

\textbf{The valid \textit{L}}.
The two conditions imply the existence of a suitable range for \textit{L}.
The first condition shows a minimum \textit{L}, and the second shows a maximum \textit{L}.
Fig.~\ref{fig:approximation_newL} shows the four learning curves with four different \textit{L} values.
The X-axis and Y-axis represent each iteration number and loss value.
When \textit{L} is $1$, the model is not learning well, and its loss is not small enough.
But it gets a smaller loss when \textit{L} is $5$.
When \textit{L} is $10$, the learning curve becomes steeper, and the final loss is smaller, indicating it learns faster and better with a larger \textit{L}. 
However, when \textit{L} is 100, the model abruptly stops learning because the 'No More Update' happens by violating the second condition.

More detailed experimental results are needed to determine the proper range of \textit{L}.
We have the two sure conditions in a mathematical form (\ref{eq:1stcond}, \ref{eq:2ndcond}) and the range of the input \textit{$p_{i}$} is set to [0, 1] since it is the probability.
The approximation function \textit{A($p_{i}$)} should be able to work with any input values when with the valid \textit{L}.
The possible \textit{$p_{i}$} values are infinite because the probability is continuous, therefore we adopt the boundary value analysis~\cite{reid1997empirical} to cover all possible \textit{$p_{i}$} values.
So the two boundaries, i.e., the two extreme values of the given range, are chosen for the test.
Even for the input values very close to 0 or 1, the approximation function should amplify them but not make them converge to exact 0 or 1.

\begin{table}[ht]
  \caption{The valid range of \textit{L}. All values between the two $p_{i}$ values are well approximated with the given \textit{L}.}
  \begin{tabular}{ccc}
    \toprule
    Accuracy & Probability & Range of \textit{L}\\
    Level \textit{t} & \textit{$p_{i}$} = \textit{t}, 1-\textit{t} & MIN $\leq$ \textit{L} $\leq$ MAX\\
    \midrule
    0.1 & 0.1, 0.9 & 5.50 $\leq$ \textit{L} $\leq$ 91.84\\
    0.01 & 0.01, 0.99 & 9.38 $\leq$ \textit{L} $\leq$ 74.97\\
    0.001 & 0.001, 0.999 & 13.85 $\leq$ \textit{L} $\leq$ 73.62\\
    $\cdots$ & $\cdots$ & $\cdots$\\
    1e-14 & 1e-14, 1-(1e-14) & 64.48 $\leq$ \textit{L} $\leq$ 73.47\\
    1e-15 & 1e-15, 1-(1e-15) & \textbf{69.08} $\leq$ \textit{L} $\leq$ \textbf{73.47}\\
    \hdashline
    1e-16 & 1e-16, 1-(1e-16) & 73.69 $\leq$ \textit{L} $\leq$ 73.47\\
  \bottomrule
\end{tabular}
\label{tab:L_range}
\end{table}

Table~\ref{tab:L_range} shows valid ranges of \textit{L} with different accuracy levels.
For instance, with the accuracy level of $0.001$ in the third row, the two extreme values $0.001$ and $0.999$ will not violate the two conditions if \textit{L} is between $13.85$ and $73.62$. 
As such, the approximation function works with any input values $[0.001, 0.999]$ when \textit{L} is in the given valid range.
If \textit{L} is selected $13.84$, then it will violate the first condition, \textit{Amplifier}, and if \textit{L} is selected $73.63$, then it will violate the second condition, \textit{No 0/1}.
The first and the second conditions regulate each the minimum and the maximum value of \textit{L}.
There is no valid range with an accuracy level of $1e^{-16}$, so the highest accuracy level is $1e^{-15}$, and we consider an integer within the range just for convenience.
The candidates are $70$, $71$, $72$, and $73$, and we chose $73$ since a larger \textit{L} is learning faster and better according to the previous observation in Fig.~\ref{fig:approximation_newL}.

\subsection{AnyLoss}
Loss functions in our method consist of the entries of a confusion matrix, which are True Negative (TN), False Negative (FN), False Positive (FP), and True Positive (TP).
Diverse loss functions can be generated since they are in a differentiable form. 
Our loss function, \textit{AnyLoss}, is defined as (\ref{eq:lossfunction}), where the \textit{f(TN,FN,FP,TP)} is a function of an evaluation metric score represented with a confusion matrix entries.
The score range is $[0, 1]$, and the form is \textit{`1-score'} so that the score can be maximized. 
To demonstrate the general availability of our method in diverse metrics, accuracy, F$_{\beta}$ scores, geometric mean, and balanced accuracy~\cite{brodersen2010balanced} are chosen.
The confusion matrix in a differentiable form and the derivatives of each loss function are provided to prove its availability in neural networks.

\begin{equation}
  AnyLoss = 1 - f(TN,FN,FP,TP)
  \label{eq:lossfunction}
\end{equation}

\textbf{Confusion Matrix}.
Our method constructs the confusion matrix with the approximated probability \textbf{YH} and the ground truth \textbf{Y}.
We assume that the approximated probability \textbf{YH} is close enough to the predicted label $\widehat{\textbf{Y}}$, and this enables the evaluation metric scores to be estimated before updating weights and used as the goal of optimization.
The four entries of the confusion matrix are described as (\ref{eq:confusionmatrix}).
This shows that the estimated entries can be approximated to the actual entries.
\begin{equation}
\begin{split}
  & TN(\textbf{Y}, \widehat{\textbf{Y}} \approx \textbf{YH}) = \sum_{i=1}^{n} (1-y_{i})(1-\widehat{y_{i}}) \approx \sum_{i=1}^{n} (1-y_{i})(1-yh_{i}) \\
  & FN(\textbf{Y}, \widehat{\textbf{Y}} \approx \textbf{YH}) = \sum_{i=1}^{n} y_{i}\cdot(1-\widehat{y_{i}}) \approx \sum_{i=1}^{n} y_{i}\cdot(1-yh_{i}) \\
  & FP(\textbf{Y}, \widehat{\textbf{Y}} \approx \textbf{YH}) = \sum_{i=1}^{n} (1-y_{i})\cdot \widehat{y_{i}} \approx \sum_{i=1}^{n} (1-y_{i})\cdot yh_{i} \\
  & TP(\textbf{Y}, \widehat{\textbf{Y}} \approx \textbf{YH}) = \sum_{i=1}^{n} y_{i}\cdot \widehat{y_{i}} \approx \sum_{i=1}^{n} y_{i}\cdot yh_{i}
\end{split}
\label{eq:confusionmatrix}
\end{equation}

\textbf{Chain Rule}.
The entries are represented with composite functions through the process described in Fig.~\ref{fig:method_mlp}. Therefore, the functional form of \textit{AnyLoss} is shown as in (\ref{eq:composite}).
The chain rule~\cite{huang2006chain} is needed as shown in (\ref{eq:chainrule}), the first term, $\frac{\partial AnyLoss} {\partial yh_{i}}$, depends on each loss function, and the rest are calculated as (\ref{eq:derivative_rest}).

\begin{equation}
  AnyLoss = f(yh_{i}(p_{i}(z_{i}(x_{i}, \textbf{W}))))
  \label{eq:composite}
\end{equation}

\begin{equation}
  \frac{\partial AnyLoss} {\partial \textbf{W}} = \frac{\partial AnyLoss} {\partial yh_{i}} \times \frac{\partial yh_{i}} {\partial p_{i}} \times \frac{\partial p_{i}} {\partial z_{i}} \times \frac{\partial z_{i}} {\partial \textbf{W}}
  \label{eq:chainrule}
\end{equation}

\begin{equation}
  \frac{\partial yh_{i}} {\partial \textbf{W}} = [L\cdot yh_{i}\cdot(1-yh_{i})] \times [p_{i}\cdot(1-p_{i})] \times [\textbf{x}_{i}]
\label{eq:derivative_rest}
\end{equation}

\textbf{Diverse Loss Functions}.
\textit{AnyLoss} indicates loss functions that can target any confusion matrix-based metric, so diverse loss functions exist according to the targeted metric as depicted in (\ref{eq:customlosses}).

\begin{equation}
\begin{split}
  &L_{A} = 1 - Accuracy = 1 - \frac{TP+TN}{TP+TN+FP+FN} \\
  &L_{F} = 1 - F_{\beta} score = 1 - \frac{(1+\beta^{2})TP}{(1+\beta^{2})TP+FP+\beta^{2}FN} \\
  &L_{G} = 1 - Geometric Mean = 1 - \sqrt{\frac{TP}{TP+FN} \times \frac{TN}{TN+FP}} \\
  &L_{B} = 1 - Balanced Acc. = 1 - \frac{1}{2}\times(\frac{TP}{TP+FN} + \frac{TN}{TN+FP})
\end{split}
\label{eq:customlosses}
\end{equation}

\textbf{Partial Derivative of AnyLoss}.
How to calculate the partial derivative of the loss function aiming at accuracy, \textit{$L_{A}$}, is addressed as an example.
Still, the calculations for other loss functions that follow the same process are explained in Appendix~\ref{App_A}.
The \textit{$L_{A}$} is represented as (\ref{eq:accuracy}) with the entries (\ref{eq:confusionmatrix}), and the first term of (\ref{eq:chainrule}) for \textit{$L_{A}$}, $\frac{\partial L_{A}} {\partial yh_{i}}$, is derived as (\ref{eq:accuracyloss}), so the partial derivative of \textit{$L_{A}$} is calculated as (\ref{eq:accuracyder}).

\begin{equation}
\begin{split}
  &L_{A} = 1 - \frac{\sum_{i=1}^{n} 1 - \sum_{i=1}^{n} y_{i} - \sum_{i=1}^{n} yh_{i} + 2(\sum_{i=1}^{n} y_{i}\cdot yh_{i})}{n}
\end{split}
\label{eq:accuracy}
\end{equation}

\begin{equation}
  \frac{\partial L_{A}} {\partial yh_{i}}= \frac{\sum_{i=1}^{n} yh'_{i} - 2(\sum_{i=1}^{n} y_{i}\cdot yh'_{i})}{n} \\
\label{eq:accuracyloss}
\end{equation}

\begin{equation}
  \frac{\partial L_{A}} {\partial \textbf{W}} = \frac{\sum_{i=1}^{n} \frac{\partial yh_{i}} {\partial \textbf{W}} - 2(\sum_{i=1}^{n} y_{i}\cdot \frac{\partial yh_{i}} {\partial \textbf{W}})}{n}
\label{eq:accuracyder}
\end{equation}

\section{Experiments}
Extensive experiments demonstrate the availability of our method and its effectiveness by suggesting improved scores against the baseline models.
In addition, the performance with imbalanced datasets and the analysis of the learning speed of our method strengthens its efficiency.
The experiments are executed in the SLP structure, which is a feed-forward network, and the MLP structure, which uses a back-propagation algorithm so that we can show that our method applies to any type of neural network.
The MLP network contains one hidden layer with two nodes, and batch normalization is used at each layer.
We set the two baselines, the first is MSE which equals the sum of squared error, the initial perceptron's loss function with the delta rule~\cite{mitchell1997machine}.
Another is the BCE, a representative loss function widely used for classification~\cite{Huang_2021_CVPR} and has a solid theoretical background in neural networks~\cite{MARCHETTI2022108913}.      
The 10-fold cross-validation is used for the robust results instead of splitting train and test data, so the scores indicate the mean of validation scores. 
In addition, only the results of accuracy, F$_{1}$ score, and balanced accuracy are shown here for conciseness, and the full results, including all metrics, are shown in Appendix~\ref{App_B}. 
We have research questions for our method as follows:
\begin{itemize}
\item Can \textit{AnyLoss} generate optimized scores for any targeted metric regardless of datasets? (\ref{4.1})
\item Is \textit{AnyLoss} more efficient with imbalanced datasets? (\ref{4.1}, \ref{4.2})
\item Is the learning speed of \textit{AnyLoss} competitive against the baselins? (\ref{4.3})
\end{itemize}

The two groups of datasets are selected to discover the research questions. 
The first group contains 102 general datasets with diverse properties to see if our method can work regardless of the datasets.
The second group contains four imbalanced datasets, and the results will show how well our method works with imbalanced datasets by providing more specific results.
As common settings, 1,000 epochs for the SLP and 100 epochs for the MLP are applied.
The learning rates and the batch sizes in the MLP are different for each loss function.
The detailed settings for experiments are included in the full results in Appendix ~\ref{App_B}.

\subsection{Performance on 102 Diverse Datasets} \label{4.1}
We conduct experiments with 102 diverse datasets and the datasets were initially collected for the imbalanced classification work ~\cite{moniz2021automated}.
The datasets are depicted as having multiple domains and diverse characteristics by the author, and their metadata is shown in Table~\ref{tab:102meta}.
The description of each dataset is shown in Table~\ref{tab:102data_appendix} in the Appendix.
They can provide the diversity required for this experiment since they have different properties regarding the dataset size, the number of features, and imbalance ratios.
The imbalance ratio, the 4th column, shows the ratio of major cases versus minor cases, which are considered positive.
The experiments with these datasets can demonstrate our method's availability in diverse datasets.

\begin{table}[ht]
  \caption{Metadata of 102 Diverse Datasets. Datasets have different properties, such as the dataset size, feature numbers, and imbalance ratios.}
  \begin{tabular}{cccc}
    \toprule
    Metrics & \# Samples & \# Features & Imb. Rat.\\
    \midrule
    mean &	1930.89 &	37.55 &	5.35:1 \\
    std &	2209.16 &	52.88 &	3.62:1 \\
    min &	250.00 &	2.00 &	1.54:1 \\
    max &	9961.00 &	299.00 & 16.43:1 \\
  \bottomrule
\end{tabular}
\label{tab:102meta}
\end{table}

Experimental Results in both SLP and MLP structures are shown in Table~\ref{tab:102table}.
The numbers indicate each winning number; for instance, for Acc, our \textit{AnyLoss} L$_{A}$ in the SLP structure achieves a better accuracy score than MSE and BCE in $69$ datasets out of a total of $102$ datasets.
For other metrics, our loss functions show larger winning numbers, meaning our loss functions perform better against baseline models in more datasets for the corresponding evaluation metric.
The specific settings for each loss function are shown in Table~\ref{tab:slp102data_appendix} and Table~\ref{tab:mlp102data_appendix} in the Appendix.

\begin{table}[ht]
  \caption{Statistical results on 102 datasets. The numbers indicate the datasets with each loss function winning with the corresponding metric.}
  \begin{tabular}{ccccccc}
    \toprule
    \multicolumn{1}{c}{Evaluation} & \multicolumn{3}{c}{SLP} & \multicolumn{3}{c}{MLP}   \\
    \cmidrule(lr){2-4} \cmidrule(lr){5-7}
     Metrics & MSE & BCE & OURS & MSE & BCE & OURS \\
    \midrule
     Acc     &  13 &  20 & \textbf{69} &  16 &  20 & \textbf{66} \\
     F-1     &  2  &  10 & \textbf{90} &  5  &  3  & \textbf{94} \\
     B-Acc   &  4  &  9  & \textbf{89} &  4  &  4  & \textbf{94} \\
    \bottomrule
  \end{tabular}
  \label{tab:102table}
\end{table}

\begin{figure*}[ht]
  \centering
  \includegraphics[width=.24\textwidth]{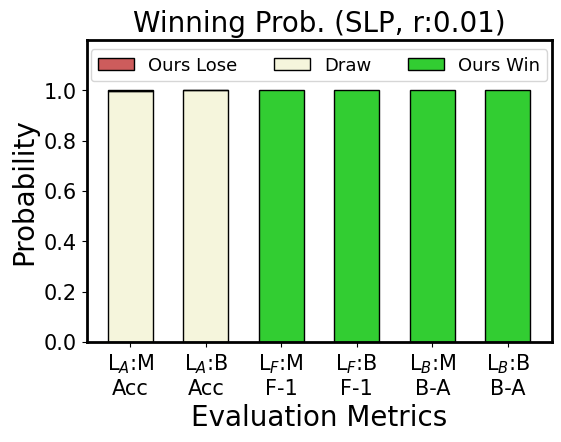}
  \includegraphics[width=.24\textwidth]{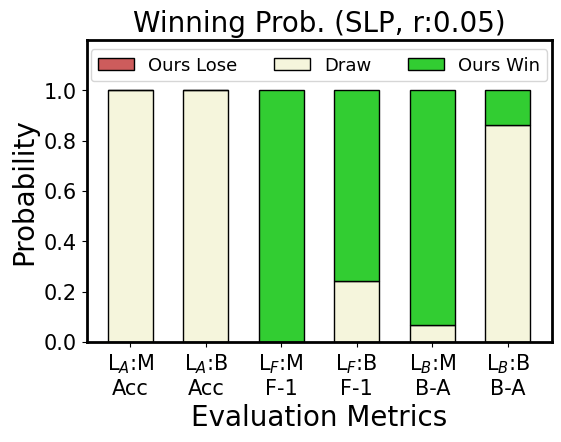}
  \includegraphics[width=.24\textwidth]{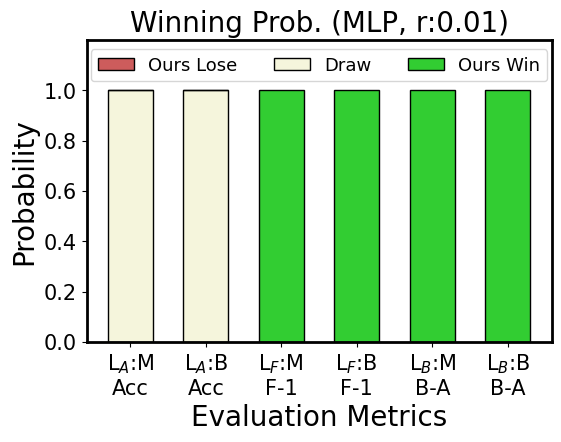}
  \includegraphics[width=.24\textwidth]{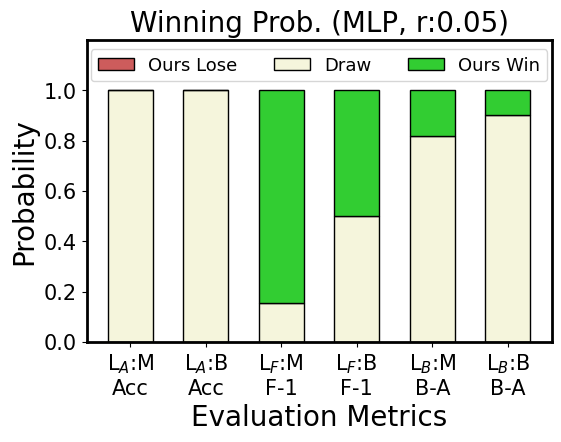}
  \caption{The winning probability of \textit{AnyLoss} against baseline models in stacked bar graphs, with the bottom for losing, the middle for drawing, and the top for winning. M and B represent, respectively, MSE and BCE. In the SLP, \textit{AnyLoss} mostly has a larger winning probability against baseline models. In some cases, a larger drawing probability is observed. In the MLP, the results are similar to those in the SLP, but more cases with a larger drawing probability are observed. And there are no red colors, meaning no cases of \textit{AnyLoss} losing.}
\label{fig:102graph_new}
\end{figure*}

\textit{AnyLoss} mostly dominates the two baselines, but it is not easy to define quantitatively their effectiveness.
Therefore, we adopt the Bayesian Sign Test~\cite{benavoli2017time}, which measures the probability that one model is better than another based on the experiment results.
This test provides the probabilities of \textit{AnyLoss} winning, drawing, or losing against each baseline model. 
The graphical results of the test are shown in Fig.~\ref{fig:102graph_new}.
The first two figures represent the results in the SLP structure, and the rest show the results in the MLP structure.
The \textit{r} represents the \textit{rope} that decides a region of practical equivalence~\cite{benavoli2017time}.
The test captures a little difference between the two models when the \textit{r} is small and distinguishes them to declare which one is better.
Reversely, when with a larger \textit{r}, the test does not distinguish the two models without a large difference, and it more likely judges them as similar in performance.
The authors in ~\cite{kruschke2015bayesian, benavoli2017time} address that it is reasonable that two models whose mean difference of scores is less than 1\% are considered practically equivalent. 
Therefore, we provide the results with \textit{r}=0.01 and \textit{r}=0.05 for further analysis.
For each metric on the X-axis, M and B indicate each MSE and BCE.
The Y-axis shows the probability of \textit{AnyLoss} winning, drawing, or losing against the corresponding baseline model.
They are stacked with the bottom for losing, the middle for drawing, and the top for winning.
The sum of the three probabilities, the height of all bars, is 1.00.
The first bar in the first figure (SLP, r:0.01) shows that \textit{AnyLoss}, L$_{A}$, has both probabilities of winning and drawing.
However, the area for drawing is much larger, which indicates L$_{A}$ has a larger probability of drawing and a little probability of winning against MSE in terms of the accuracy score.
In the second figure, when with a larger \textit{r}=0.05, \textit{AnyLoss} generally has a larger drawing probability compared to when \textit{r}=0.01 since a little difference is ignored.
In the SLP structure, our loss functions mostly have a larger winning probability against the two baseline models except for accuracy.
The results in the MLP structure are very similar to those in the SLP.
The red color is not observed in any graph, which means that the probability of MSE or BCE being better than \textit{AnyLoss} is 0 for all metrics.
Consequently, \textit{AnyLoss} shows mostly better performance of all metrics except for accuracy on 102 general datasets.
Considering the objective of the two baselines, pursuing higher accuracy, our method shows good performance, which was our intention.
The numerical results of the test are shown in Table~\ref{tab:slp102data_appendix} and Table~\ref{tab:mlp102data_appendix} in the Appendix.

To observe if our method works better with imbalanced datasets, we classify 102 datasets into four groups based on their imbalance ratios.
Whether better or not has been shown in the previous results, now we provide how much better \textit{AnyLoss} against each baseline model in Table~\ref{tab:102imbalance}.
The symbol $\Delta$ means the difference, therefore the results indicate `\textit{AnyLoss} score - baseline score'.
For example, the 31 datasets in the first group have imbalance ratios between 60:40 and 70:30, and \textit{AnyLoss} accuracy score is better than MSE as 0.017 of mean $\pm$ 0.035 of standard deviation.
Accuracies show irregular patterns. However, there is a tendency for the F-1 score and balanced accuracy, a larger difference in the group of highly imbalanced datasets.
From top to bottom, little imbalanced to highly imbalanced, differences become larger, which means our method works better with more imbalanced datasets.
This result shows that \textit{AnyLoss} can be more suitable than the two baseline models in dealing with highly imbalanced datasets.  

\begin{table}[ht]
  \caption{The effect of our method in different groups based on imbalance ratios. The symbol $\Delta$ indicates the difference, meaning how much our method is better than each model. The numbers mean `\textit{AnyLoss} score - baseline score'. From top to bottom, datasets are more imbalanced, and larger scores are observed, which indicates our method performs better in more imbalanced datasets.}
  \begin{tabular}{ccccc}
    \toprule
     \multicolumn{1}{c}{Eva.} & \multicolumn{2}{c}{SLP} & \multicolumn{2}{c}{MLP}  \\
     \cmidrule(lr){2-3} \cmidrule(lr){4-5}
     \multicolumn{1}{c}{Met.} & $\Delta$ MSE &$\Delta$ BCE & $\Delta$ MSE & $\Delta$ BCE \\
    \midrule
      & \multicolumn{4}{l}{\textit{Imbalance Ratio: 60:40 $\sim$ 70:30 | 31 Datasets}} \\
     Acc     &    0.017$\pm$0.035   &   0.007$\pm$0.029   &    0.004$\pm$0.025 &    0.009$\pm$0.016 \\
     F-1     &    0.081$\pm$0.095   &   0.057$\pm$0.092   &    0.040$\pm$0.100 &    0.040$\pm$0.091 \\
     B-A     &    0.043$\pm$0.052   &   0.026$\pm$0.049   &    0.007$\pm$0.046 &    0.009$\pm$0.044 \\
    \midrule
      & \multicolumn{4}{l}{\textit{Imbalance Ratio: 70:30 $\sim$ 80:20 | 19 Datasets}} \\
     Acc     &    0.021$\pm$0.029   &   0.017$\pm$0.027   &    -0.058$\pm$0.207 &    -0.062$\pm$0.207 \\
     F-1     &    0.186$\pm$0.156   &   0.142$\pm$0.132   &    0.114$\pm$0.117 &    0.096$\pm$0.124 \\
     B-A     &    0.086$\pm$0.065   &   0.067$\pm$0.057   &    0.067$\pm$0.061 &    0.055$\pm$0.056 \\
    \midrule
      & \multicolumn{4}{l}{\textit{Imbalance Ratio: 80:20 $\sim$ 90:10 | 37 Datasets}} \\
     Acc     &    0.008$\pm$0.021   &   0.002$\pm$0.009   &    0.002$\pm$0.006 &    0.002$\pm$0.007 \\
     F-1     &    0.208$\pm$0.209   &   0.122$\pm$0.131   &    0.138$\pm$0.144 &    0.137$\pm$0.151 \\
     B-A     &    0.128$\pm$0.120   &   0.084$\pm$0.086   &    0.084$\pm$0.080 &    0.079$\pm$0.082 \\
    \midrule
      & \multicolumn{4}{l}{\textit{Imbalance Ratio: 90:10 $\sim$  | 15 Datasets}} \\
     Acc     &    0.004$\pm$0.008   &   0.001$\pm$0.005   &    0.002$\pm$0.009 &    0.004$\pm$0.009 \\
     F-1     &    0.304$\pm$0.144   &   0.167$\pm$0.127   &    0.253$\pm$0.179 &    0.222$\pm$0.127 \\
     B-A     &    0.226$\pm$0.097   &   0.170$\pm$0.088   &    0.152$\pm$0.076 &    0.128$\pm$0.076 \\         
  \bottomrule
  \multicolumn{5}{l}{score: mean $\pm$ standard deviation}
\end{tabular}
\label{tab:102imbalance}
\end{table}


\subsection{Performance on 4 Imbalanced Datasets} \label{4.2}
We conduct experiments with 4imbalanced datasets to observe our method's availability with imbalanced datasets, and their descriptions are shown in Table~\ref{tab:4data}.
The first dataset is generated with the Scikit\-learn library \textit{datasets.make\_classification}, and others are collected from \textit{Kaggle}.
For \textit{Credit Card}~\cite{credit} and \textit{Breast Cancer}~\cite{breast} datasets, we intentionally make them more imbalanced using the Python library \textit{sample}.
In addition, we adopt one of the SOTA surrogate approaches, the Score-Oriented Loss (SOL) ~\cite{MARCHETTI2022108913} for comparison.
SOL proposes a similar approach to ours, constructing a confusion matrix to generate any metric, therefore we compare it with ours.

\begin{table}[ht]
  \caption{Description of 4 Imbalanced Datasets}
  \begin{tabular}{lccc}
    \toprule
    Datasets & \# Samples & \# Features & Imb. Rat.\\
    \midrule
    \#1. Random & 10,000 & 2 & 9:1 \\   
    \#2. Credit Card~\cite{credit} & 298,531 & 29 & 20:1 \\     
    \#3. Breast Cancer~\cite{breast} & 393 & 30 & 10:1  \\   
    \#4. Diabetes~\cite{diabetes} & 100,000 & 8 & 11:1 \\    
  \bottomrule
\end{tabular}
\label{tab:4data}
\end{table}

Experimental Results in both SLP and MLP structures are shown in Table~\ref{tab:4table}.
The table provides only one corresponding metric that each \textit{AnyLoss} aims at.
For example, L$_{A}$ is made for higher accuracy, so we only present the accuracy score of it.
In the first row, $0.922$ is the accuracy score of L$_{A}$, and $0.632$ in the second row is the F-1 score of L$_{F_{1}}$.
The scores of our \textit{AnyLoss} are similar to SOL in the SLP and mostly better than SOL in the MLP structure but the differences are not large.
Baseline models generally show good performance only for accuracy and large differences for other metrics.
The full experiment results including other metric scores of each \textit{AnyLoss} are shown in Table~\ref{tab:slp4data_appendix} and Table~\ref{tab:mlp4data_appendix} in the Appendix.
The results in Table~\ref{tab:4table} show that it is possible to get an optimized score of the desired metric with \textit{AnyLoss} in imbalanced datasets. 

\begin{table}[ht]
\small
  \caption{Scores on four imbalanced datasets. \textit{AnyLoss} and SOL show similar performance and are better than the two baseline models. In particular, large differences are observed in metrics other than accuracy.}
  \begin{tabular}{ccccccccc}
    \toprule
    \multicolumn{1}{c}{Eva} & \multicolumn{4}{c}{SLP} & \multicolumn{4}{c}{MLP}  \\
    \cmidrule(lr){2-5} \cmidrule(lr){6-9}
     Met & MSE & BCE & SOL & OUR & MSE & BCE & SOL & OUR \\
    \midrule
     &\multicolumn{8}{l}{\textit{\#1. Random}} \\
     Acc     &  0.898 &  0.921 & \textbf{0.922} & \textbf{0.922}  &  0.920 &  0.924 &  0.923  & \textbf{0.925} \\
     F-1     &  0.075 &  0.546 & \textbf{0.634} &  0.632  &  0.527 &  0.590 &  0.636  & \textbf{0.637} \\
     B-A   &  0.519 &  0.714 & \textbf{0.875} &  \textbf{0.875}   &  0.711 &  0.748 &  0.858  & \textbf{0.862}\\
    \midrule
    &\multicolumn{8}{l}{\textit{\#2. Credit Card}} \\
     Acc     &  0.989 &  0.990 & \textbf{0.991} &  0.990 &  0.991 & \textbf{0.994} &  0.992 & \textbf{0.994} \\
     F-1     &  0.880 &  0.884 & 0.896 &  \textbf{0.903} &  0.889 &  0.929 &  0.944 & \textbf{0.946}\\
     B-A   &  0.894 &  0.897 & \textbf{0.957} &  \textbf{0.957}  &  0.904 &  0.942 &  0.950  &\textbf{0.957}\\
    \midrule
     &\multicolumn{8}{l}{\textit{\#3. Breast Cancer}} \\
     Acc     &   0.979 &  0.984 & 0.980 &  \textbf{0.987}  &  0.982 &  \textbf{0.995} &  0.985 & \textbf{0.995} \\
     F-1     &   0.863 &  0.901 & \textbf{0.922} &  0.901  &  0.833 &  0.966 &  0.955 & \textbf{1.000} \\
     B-A   &   0.887 &  0.927 & \textbf{0.965} &  0.963 &  0.900 &  0.971 &  0.975    & \textbf{0.983}\\
    \midrule
    & \multicolumn{8}{l}{\textit{\#4. Diabetes}} \\
     Acc     &   0.939 &  0.959 & 0.898 &  \textbf{0.960}  &  0.956 &  0.960 &  \textbf{0.961} & \textbf{0.961} \\
     F-1     &   0.453 &  0.710 & 0.670 &  \textbf{0.732}  &  0.655 &  0.717 &  \textbf{0.736 }& \textbf{0.736} \\
     B-A   & 0.647 &  0.790 & 0.880 &  \textbf{0.885}  &  0.766 &  0.793 &  0.865 & \textbf{0.877} \\            
  \bottomrule
\end{tabular}
\label{tab:4table}
\end{table}

Table~\ref{tab:4resample} compares our method with resampling strategies on the four datasets in the SLP structure.
The resampling strategies are applied only to BCE since it is better than MSE in the previous experiments.
BSO indicates BCE with SMOTE~\cite{chawla2002smote}, and BRU means BCE with the random under-sampling.
As for a resampling rate, we try multiple (major : minor) rates from (1:1) to (1:0.1) with 0.1 intervals, and the result with the highest F-1 score is chosen from the results from all different rates.
Over-sampling and under-sampling strategies show improved results w.r.t. basic BCE, with scores similar to those of \textit{AnyLoss}.
On the other hand, our method does not distort data distribution and achieves a goal at once without multiple times of experiments to choose the best.

\begin{table}[ht]
  \caption{Comparison with resampling strategies. BSO and BRU indicate BCE with SMOTE and BCE with random under-sampling. Both strategies improve performance, but \textit{AnyLoss} still shows competitive results.}
  \begin{tabular}{ccccccc}
    \toprule
    \multicolumn{1}{c}{Eva.} & \multicolumn{6}{c}{SLP}  \\
     Met. & BSO & BRU & OURS & BSO & BRU & OURS \\
    \midrule
     &\multicolumn{3}{l}{\textit{\#1. Random}} & \multicolumn{3}{l}{\textit{\#3. Breast Cancer}} \\
     Acc     &  0.915 &  0.916 & \textbf{0.922} &  \textbf{0.987} &  0.984 & \textbf{0.987}  \\
     F-1     &  \textbf{0.637} &  \textbf{0.637} & 0.632 &  \textbf{0.930} &  0.914 & 0.901 \\
     B-A     &  0.823 &  0.821 & \textbf{0.875} &  \textbf{0.970} &  0.965 & 0.963 \\
    \midrule
    &\multicolumn{3}{l}{\textit{\#2. Credit Card}} & \multicolumn{3}{l}{\textit{\#4. Diabetes}} \\
     Acc     &  \textbf{0.992} &  \textbf{0.992} & 0.990  &  0.959 &  0.959 & \textbf{0.960} \\
     F-1     &  0.912 &  \textbf{0.913} & 0.903  &  0.727 &  0.727 & \textbf{0.732} \\
     B-A     &  0.932 &  0.932 & \textbf{0.957} &  0.809 &  0.809 & \textbf{0.885} \\
  \bottomrule
\end{tabular}
\label{tab:4resample}
\end{table}

\subsection{Learning Time} \label{4.3}
We measure the learning time of \textit{AnyLoss}, SOL, and the two baseline models in the same epochs, and this shows the pure learning time per epoch, i.e., learning speed.
We also analyze the loss curve to conjecture the required epochs for each method.
This gives us inspiration about the practical learning time of each method.

\textbf{Learning Speed}. 
Under the same number of epochs, we measure the learning time of \textit{AnyLoss}, SOL, and the two baseline models with the four imbalanced datasets in both SLP and MLP structures.
The same condition is required for all loss functions, so the same settings, learning rates, and batch sizes are applied instead of the already used settings.
The results are as Table~\ref{tab:learningtime}.
\begin{table}[ht]
  \caption{Ratios of learning time based on BCE's learning time with the four datasets. L$_{mean}$ and S$_{mean}$ mean each the average ratio of all \textit{Anyloss} and SOL. In both SLP and MLP structures, MSE is faster than BCE, \textit{AnyLoss} is similar to BCE, and the SOL is the slowest.}
  \begin{tabular}{ccccccc}
    \toprule
     \multicolumn{1}{c}{Data} & \multicolumn{3}{c}{SLP} & \multicolumn{3}{c}{MLP}   \\
     \cmidrule(lr){2-4} \cmidrule(lr){5-7}
     sets    &   MSE   & S$_{mean}$ & L$_{mean}$&   MSE   & S$_{mean}$ & L$_{mean}$ \\
    \midrule
     \#1     &  1.001  &   1.120   &   1.005   &  0.888  &   1.158  &   1.031 \\
    \midrule
     \#2     &  0.893  &   1.175   &   0.968   &  0.783  &   1.090  &   0.993 \\
    \midrule
     \#3     &  0.821  &   1.103   &   0.901   &  0.994  &   1.156  &   1.011 \\
    \midrule
     \#4     &  0.905  &   1.151   &   1.041   &  0.804  &   1.082  &   0.987 \\
  \bottomrule
  \multicolumn{7}{l}{The standard is BCE's learning time, i.e., BCE: 1.00.}
\end{tabular}
\label{tab:learningtime}
\end{table}
The ratios are calculated based on the learning time of BCE, therefore the ratios in the table are calculated as its learning time divided by the learning time of BCE.
The differences among learning times of all \textit{AnyLoss} are small, so we used their average value, L$_{mean}$.
In the same sense, S$_{mean}$ indicates the average value of all SOL functions.
The ratios are learning speed because the result times are based on the same number of epochs.
The results are similar in both the SLP and MLP structures.
MSE is faster than BCE, \textit{AnyLoss} is similar to BCE, and SOL is the slowest.
Our method shows a competitive learning speed against BCE since our approach includes one more step, the approximation step.
Its calculation is simple, so it rarely affects the learning time.
On the other hand, SOL shows slower speed due to its complicated process.
This result shows the competitiveness of our method in terms of pure learning speed in neural networks.

\begin{figure}[ht]
  \centering
  \includegraphics[width=.23\textwidth]{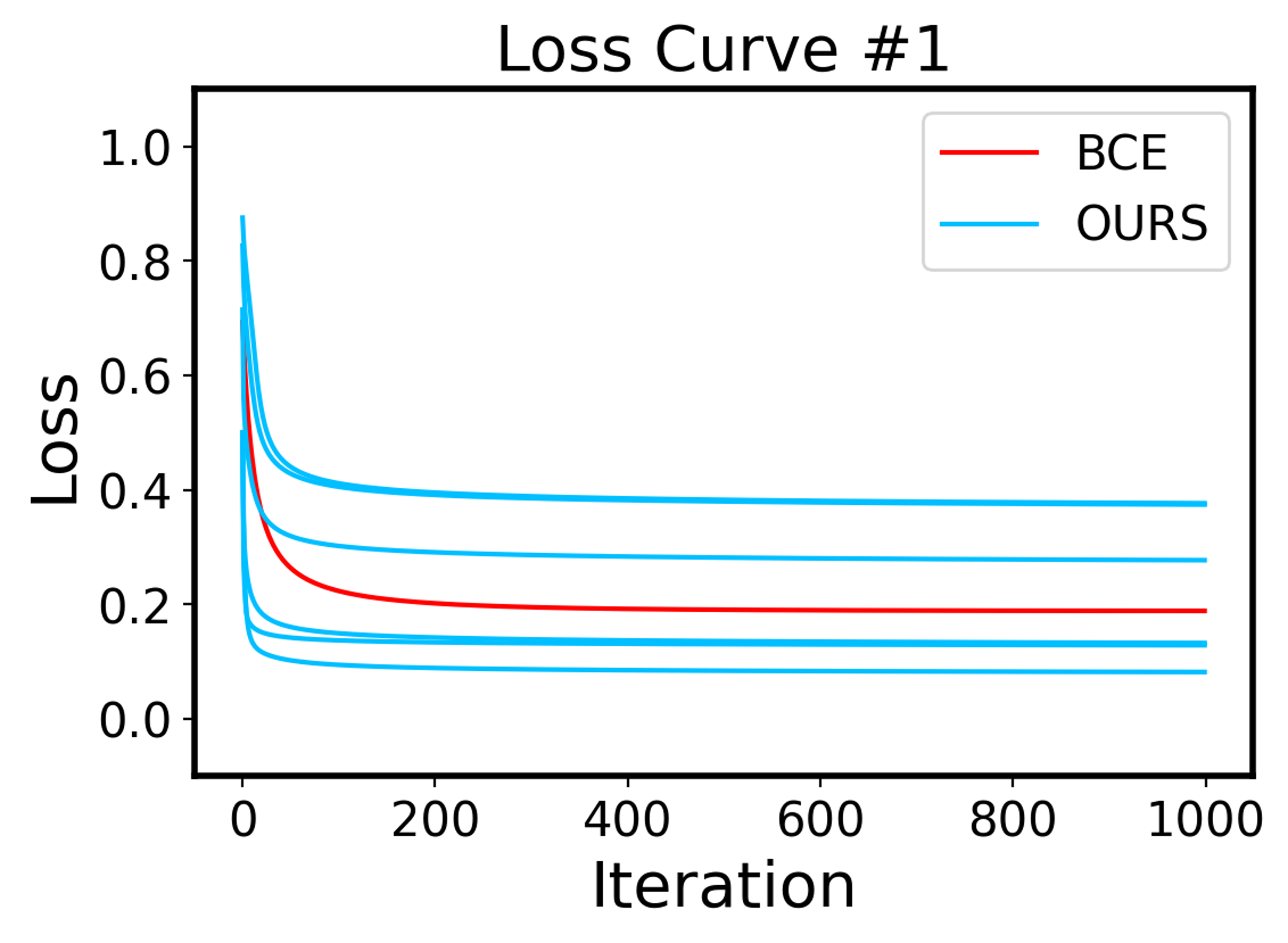}\hfill
  \includegraphics[width=.23\textwidth]{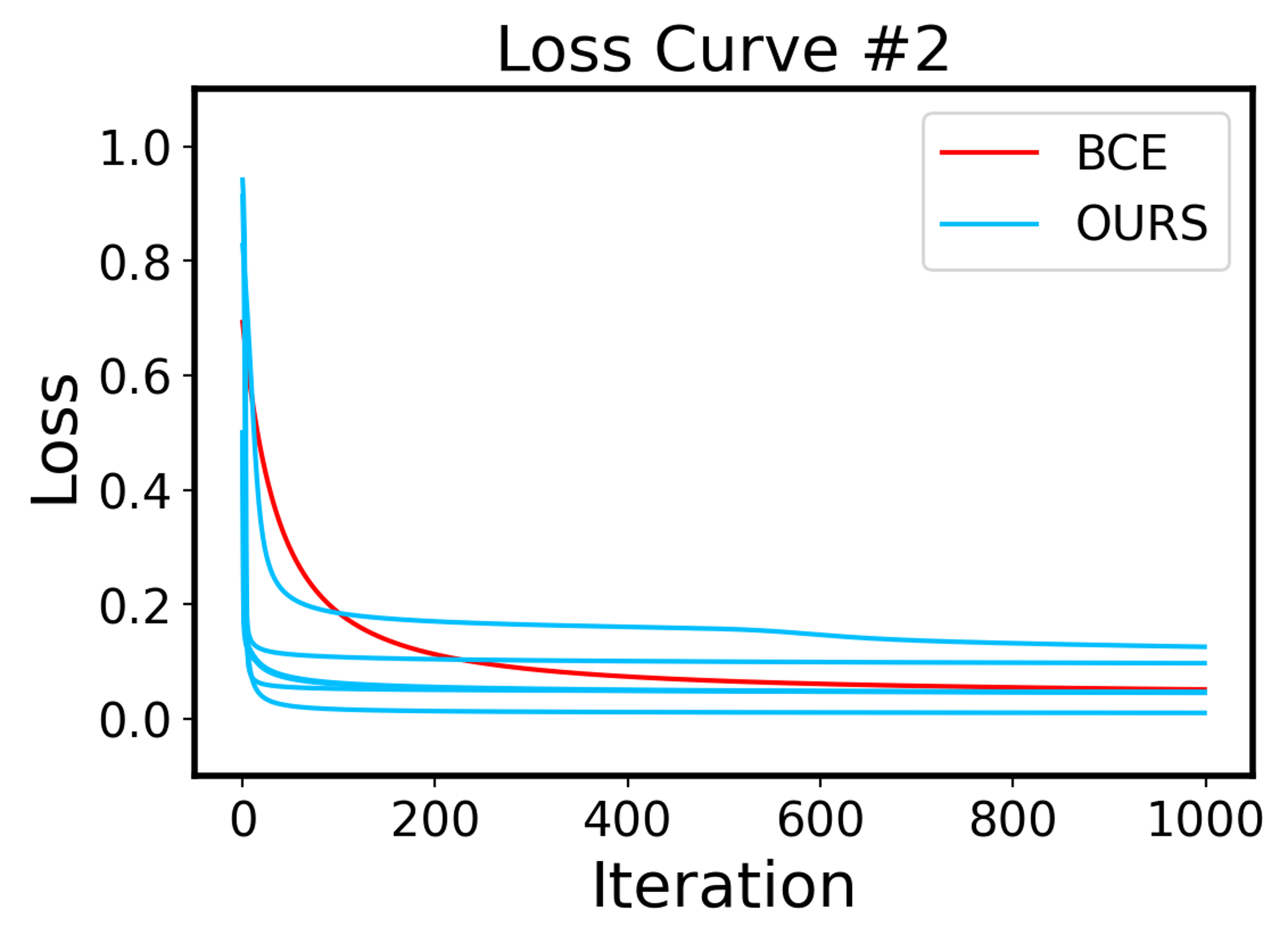}
  \includegraphics[width=.23\textwidth]{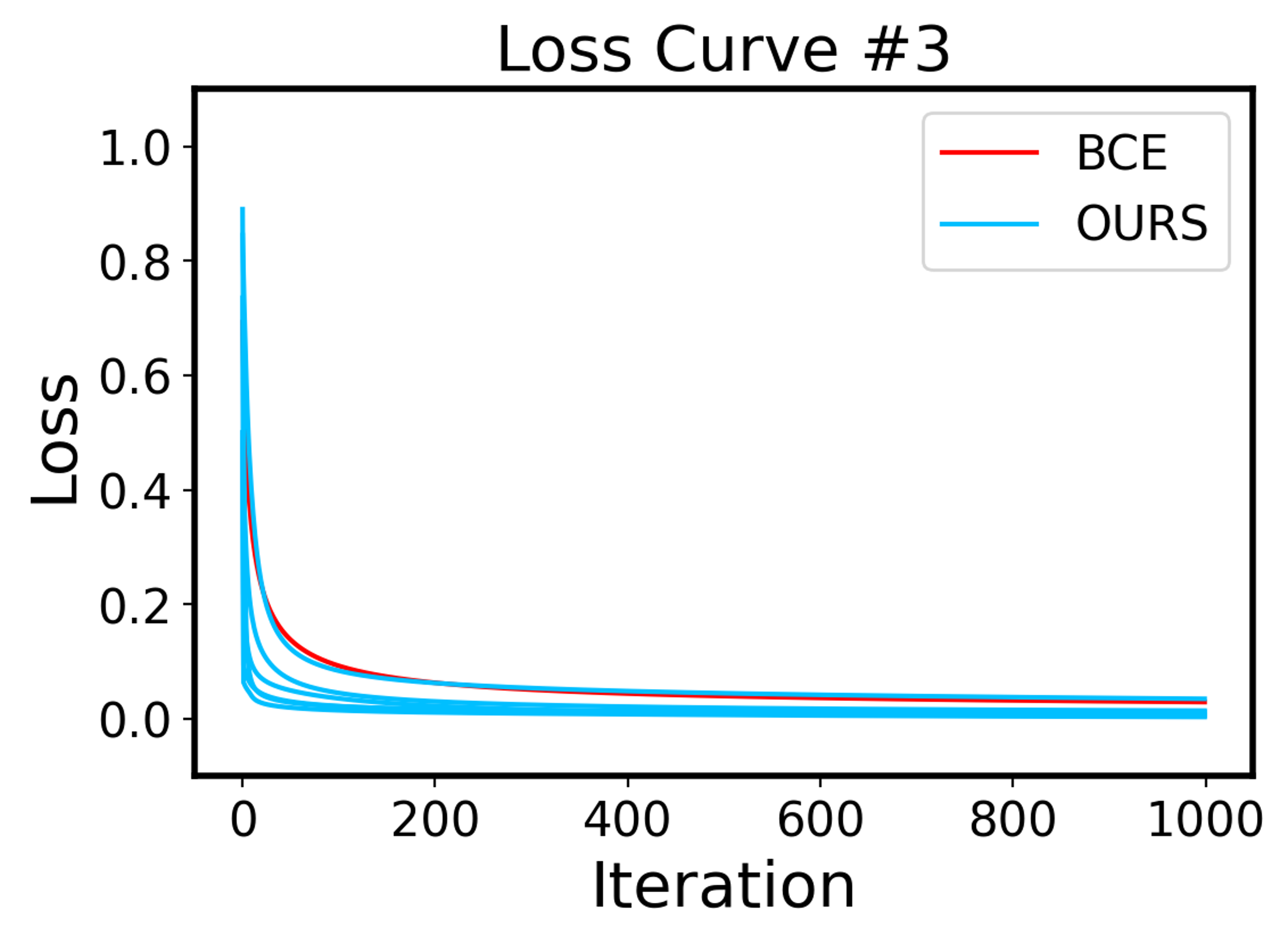}\hfill
  \includegraphics[width=.23\textwidth]{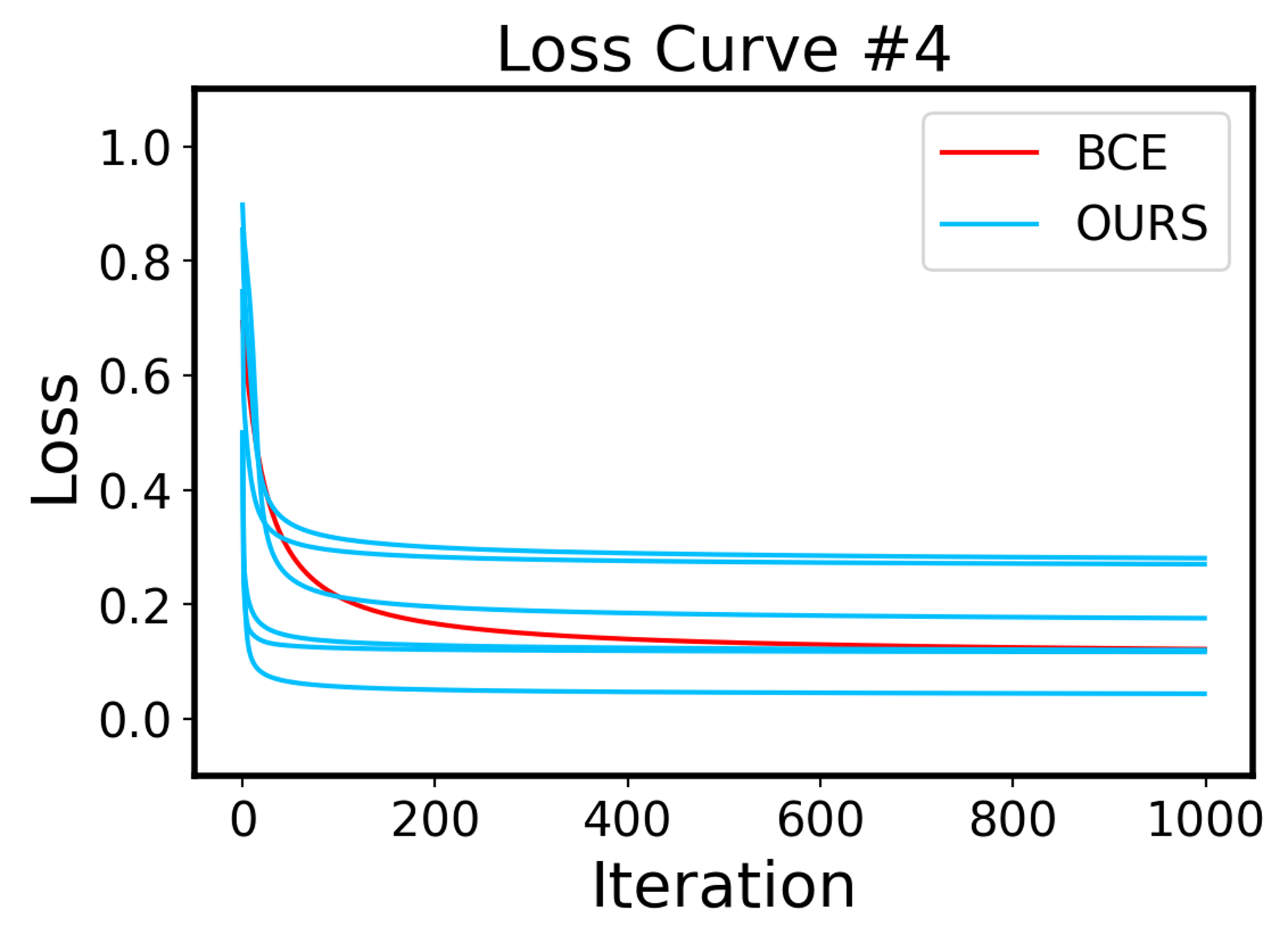}
  \caption{The learning curves BCE vs. \textit{AnyLoss}. They show similar slopes in dataset 1, and \textit{AnyLoss} shows a steeper slope in other datasets, meaning it learns faster and needs fewer epochs to achieve its optimal point.}
\label{fig:learningtime_epoch}
\end{figure}

\textbf{Practical Learning Time}.
In reality, the more important figure about learning time is the practically required time for learning, not pure speed.
We measured the pure learning time in the previous step; however, the amount of time needed for learning is not explained by the unit learning speed because it largely depends on the required epochs.
Therefore, we analyze the loss learning curves to discover each method's practical required number of epochs.
We choose BCE and \textit{AnyLoss}, the fastest learning speed for this experiment.
Different learning rates for each method have to be applied since they decide the slope of the curve and performance.
For example, if we increase the learning rate for a certain loss function, it may learn faster, but it does not guarantee the best performance.
We already have the appropriate learning rates used for the previous experiments, so they are used for this experiment.
Fig.~\ref{fig:learningtime_epoch} shows how fast each method learns and converges to its minimum loss value.
The X-axis and Y-axis represent each the number of epochs and loss value.
The red line represents BCE's learning curve, and the blue lines show all \textit{AnyLoss}, i.e. L$_{A}$, L$_{F1}$, etc.
The minimum loss value of each function is different since they have their own goal defined by different loss functions.
The first graph, `Loss Curve \#1' shows the results on the first dataset, and the slopes of all lines look similar. 
Before the 100 epochs, they seem to have mostly achieved the minimum values.
However, in the other three graphs, the slopes of the blue lines look steeper than those of BCE.
For a more detailed analysis, we calculate the achievement rate, described as (\ref{eq:achievementrate}).
\begin{equation}
  Achievement Rate = \frac{Initial Loss - Current Loss} {Initial Loss - Final Loss}
  \label{eq:achievementrate}
\end{equation}
This rate indicates how much loss decreased by a certain period of epochs.
For instance, if the initial loss value is $1.0$, the loss value is $0.7$ when the epoch is $100$, and the final loss value is $0.4$, then it decreases a total of $0.6$ for a whole period of epochs.
It decreases by $0.3$ for the first $100$ epochs. Therefore, the achievement rate at epoch $100$ is $0.5$, and the rate at the last epoch is always $1.0$.
Table~\ref{tab:learningtime_epoch} shows the results of the achievement rates, and L$_{mean}$ is the average of the achievement rates of all \textit{AnyLoss}.
In the first dataset, the two achievement rates are similar, and the difference is not large.
In other datasets, the average achievement rate of \textit{AnyLoss} is larger than BCE at all epochs, which indicates \textit{AnyLoss} generally learns faster than BCE. Consequently, \textit{AnyLoss} normally needs a smaller number of epochs leading to a shorter learning time.

\begin{table}[ht]
  \caption{Achievement rates on four datasets}
  \begin{tabular}{cccccccc}
    \toprule
    Data     &    Loss      & \multicolumn{6}{c}{Epoch}   \\
    sets     &    Fun.      &   100   &   200    &    300   &   400   &    500   &    1,000 \\
    \midrule
     \#1     &  BCE         &  0.930  &   0.974   &  0.987  &  0.993  &   0.996  &   1.0   \\
             &  L$_{mean}$  &  0.951  &   0.973   &  0.982  &  0.987  &   0.991  &   1.0   \\
    \midrule
     \#2     &  BCE         &  0.790  &   0.904   &  0.945  &  0.965  &   0.977  &   1.0   \\
             &  L$_{mean}$  &  0.970  &   0.980   &  0.984  &  0.987  &   0.989  &   1.0   \\
    \midrule
     \#3     &  BCE         &  0.904  &   0.949   &  0.967  &  0.978  &   0.984  &   1.0   \\
             &  L$_{mean}$  &  0.961  &   0.978   &  0.985  &  0.990  &   0.993  &   1.0   \\
    \midrule
     \#4     &  BCE         &  0.838  &   0.922   &  0.953  &  0.969  &   0.979  &   1.0   \\
             &  L$_{mean}$  &  0.959  &   0.978   &  0.985  &  0.990  &   0.993  &   1.0   \\
  \bottomrule
\end{tabular}
\label{tab:learningtime_epoch}
\end{table}

\section{Additional Experimental Results}
To reinforce our argument on the availability of \textit{AnyLoss}, we provide more experimental results with an advanced MLP architecture and a larger dataset. In addition, we present the performance change results along with the change of the amplifying scale \textit{L} to support our approach suggested in the previous section to determine a good \textit{L} value of the approximation function. 

\subsection{Advanced Architecture}
The SLP and MLP structures were chosen for the analysis from the perspective of neural networks, but they may not be enough to prove the effectiveness of our method convincingly. Therefore, we provide more experimental results with an advanced MLP architecture. We did an image classification task with the ResNet50~\cite{he2016deep} and the MNIST\_784~\cite{mnist_784} dataset. The dataset originally had 10 class labels, so we chose one class label as a positive case to make a binary classification task and others were given a negative label. The results support that \textit{AnyLoss} still works with more complex architecture as shown in Table~\ref{tab:resnet}.

\begin{table}[ht]
  \caption{Scores of different loss functions in the ResNet50 architecture with the MNIST\_784 dataset. \textit{AnyLoss} shows better performance in each evaluation metric.}
  \begin{tabular}{cccc}
    \toprule
     Metrics & MSE & BCE & OURS \\
    \midrule
     Acc     &  0.994 &  0.994 & \textbf{0.998}  \\
     F-1     &  0.965 &  0.965 & \textbf{0.980}  \\
     B-A     &  0.972 &  0.970 & \textbf{0.980}  \\
  \bottomrule
\end{tabular}
\label{tab:resnet}
\end{table}

\subsection{Large Dataset}
In terms of dataset size, a new loss function should properly work with any size of datasets, therefore, we provide experiment results with a large dataset, \textit{Huge Titanic}~\cite{huge_titanic} (\#samples: 799,603, \#features: 7, imbalance ratio: 0.62:0.38). The original number of samples is a million but the dataset contains many missing data, so we used 799,603 after pre-processing. Table~\ref{tab:titanic} shows that \textit{AnyLoss} works well in both structures with any metrics.

\begin{table}[ht]
  \caption{Scores on the Huge Titanic dataset. \textit{AnyLoss} is better than the two baseline models in both SLP and MLP structures.}
  \begin{tabular}{ccccccc}
    \toprule
    \multicolumn{1}{c}{Eva} & \multicolumn{3}{c}{SLP} & \multicolumn{3}{c}{MLP}  \\
    \cmidrule(lr){2-4} \cmidrule(lr){5-7}
     Met & MSE & BCE & OURS & MSE & BCE & OURS \\
    \midrule
     Acc     &  0.800 &  0.806 & \textbf{0.814}  &  0.807 &  0.814 & \textbf{0.822} \\
     F-1     &  0.726 &  0.734 & \textbf{0.739}   &  0.730 &  0.729 & \textbf{0.746} \\
     B-A   &  0.780 &  0.786 & \textbf{0.788}    &  0.783  &  0.784 & \textbf{0.796} \\     
  \bottomrule
\end{tabular}
\label{tab:titanic}
\end{table}

\subsection{Determination of L}
Determination of the amplifying scale \textit{L} is essential for the approximation function since it decides how to amplify the given probability from the Sigmoid function. As discussed earlier, an invalid value of \textit{L} can cause malfunctioning of the approximation function, and then it will not provide the desired results even if it looks working properly. So, we executed experiments to observe how performances change along with the change of the amplifying scale \textit{L}. The experiment results in the MLP structure with different \textit{L} values on different datasets are shown in Table~\ref{tab:detL4} and Table~\ref{tab:detL102}. We used 5 different \textit{L} values including 73 which has been used as a value of \textit{L} in this paper. The 3 evaluation metrics are calculated on the 4 imbalanced datasets and 102 diverse datasets as done earlier. In the results, the value of 73 shows better performance than others but the difference is not large. Therefore, this topic, determining a good L, can be considered as future work.

\begin{table}[ht]
  \caption{Scores on 4 imbalanced datasets with different L. The value 73 which we chose earlier mostly shows the best score in each dataset and evaluation metric even though big differences are not observed.}
  \begin{tabular}{ccccccc}
    \toprule
     Dataset & Metric & L=30 & L=50 & L=73 & L=90 & L=110 \\
    \midrule
     Random& Acc     &  0.915 &  0.923 & \textbf{0.924} & 0.895  &  0.922 \\
     Generated& F-1     &  0.636 &  0.632 & \textbf{0.637} & 0.634  &  0.635 \\
     & B-A     &  0.840 &  0.834 & \textbf{0.852} & 0.826  &  0.825 \\
    \midrule
     Credit & Acc     &  0.952 &  0.994 & \textbf{0.995} &  0.994 &  0.994 \\
     Card & F-1     &  0.935 &  0.942 & \textbf{0.945} &  \textbf{0.945} &  0.944 \\
      & B-A     &  0.950 &  0.915 & \textbf{0.962} &  0.956 &  0.871 \\
    \midrule
     Breast & Acc     &  0.998 &  0.995 & \textbf{1.000} &  0.997 &  0.997 \\
     Cancer & F-1     &  0.980 &  0.980 & \textbf{0.989} &  0.980 &  0.969 \\
      & B-A     &  0.949 &  0.950 & \textbf{1.000} &  0.989 &  0.996 \\
    \midrule
    Diabetes & Acc     &  0.948 &  \textbf{0.961} & \textbf{0.961} &  \textbf{0.961}  &  \textbf{0.961} \\
    Prediction &  F-1     &  0.728 &  0.733 & \textbf{0.736} &  0.735 &  0.734 \\
      & B-A     &  0.837 &  0.859 & \textbf{0.882} &  0.832 &  0.826 \\            
  \bottomrule
\end{tabular}
\label{tab:detL4}
\end{table}

\begin{table}[ht]
  \caption{Scores on 102 diverse datasets with different L. Each number in the table indicates the winning number, i.e., a model with L=73 shows the best score in 31 out of 102 datasets. The value 73 shows the best winning number.}
  \begin{tabular}{cccccc}
    \toprule
      Metric & L=30 & L=50 & L=73 & L=90 & L=110 \\
    \midrule
      Acc     &  21 &  18 & \textbf{31} & 21  &  24 \\
      F-1     &  21 &  19 & \textbf{27} & 21  &  16 \\
      B-A     &  17 &  20 & \textbf{26} & 17  &  23 \\
  \bottomrule
\end{tabular}
\label{tab:detL102}
\end{table}


\section{Conclusion}
In this paper, we introduce \textit{AnyLoss}, which can aim at a specific confusion matrix-based evaluation metric with the approximation function.
Our method enables any confusion matrix-based metric to be directly set as a goal of learning, i.e., loss function, in a neural network architecture to achieve a unique goal of the binary classification task.
With mathematical approaches, we prove that \textit{AnyLoss} are differentiable and provide their derivatives.
Analysis of the approximation function provided with experiments helps users use it more efficiently in practice.
Extensive experiment results demonstrate that our method achieves a goal in both SLP and MLP structures with diverse datasets while maintaining a competitive learning time.
It, especially, shows considerable performance in imbalanced datasets although it is available with any type of dataset.
A similar approach to handling multi-class classification tasks or determining a good value of \textit{L} to improve its efficiency can be interesting research topics for future research.
All the datasets and codes used for the experiments are available at https://github.com/doheonhan/anyloss.

%
\bibliographystyle{ACM-Reference-Format}
\bibliography{doheon}

\appendix

\section{Calculation of Partial Derivatives} \label{App_A}

\subsection{F-beta scores}
The loss function aiming at F-$\beta$, \textit{$L_{F}$}, is represented as (\ref{eq:fbeta_appendix}) with the confusion matrix (\ref{eq:confusionmatrix}).
The first term of (\ref{eq:chainrule}) for \textit{$L_{F}$} is as (\ref{eq:fbetaloss_appendix}).
The partial derivative of \textit{$L_{F}$} is as (\ref{eq:fbetader}).

\begin{equation}
  L_{F} = 1 - \frac{(1+\beta^{2})(\sum_{i=1}^{n} y_{i} \cdot yh_{i})}{\beta^{2}\sum_{i=1}^{n} y_{i} + \sum_{i=1}^{n} yh_{i}}
\label{eq:fbeta_appendix}
\end{equation}

\begin{equation}
\begin{split}
  & \frac{\partial L_{F}} {\partial yh_{i}} = -(1+\beta^{2}) \times \\ 
  & \resizebox{0.9\hsize}{!}{$\frac{(\sum_{i=1}^{n} y_{i} \cdot yh'_{i})(\beta^{2}\sum_{i=1}^{n} y_{i} + \sum_{i=1}^{n} yh_{i}) - (\sum_{i=1}^{n} yh'_{i})(\sum_{i=1}^{n} y_{i} \cdot yh_{i})}{(\beta^{2}\sum_{i=1}^{n} y_{i} + \sum_{i=1}^{n} yh_{i})^2}$} \\
\end{split}
\label{eq:fbetaloss_appendix}
\end{equation}

\begin{equation}
\begin{split}
  & \frac{\partial L_{F}} {\partial \textbf{W}} = -(1+\beta^{2}) \times \\
  & \resizebox{0.9\hsize}{!}{$\frac{(\sum_{i=1}^{n} y_{i} \cdot \frac{dyh_{i}} {d\textbf{W}})(\beta^{2}\sum_{i=1}^{n} y_{i} + \sum_{i=1}^{n} yh_{i}) - (\sum_{i=1}^{n} \frac{dyh_{i}} {d\textbf{W}})(\sum_{i=1}^{n} y_{i} \cdot yh_{i})}{(\beta^{2}\sum_{i=1}^{n} y_{i} + \sum_{i=1}^{n} yh_{i})^2}$}
\end{split}
\label{eq:fbetader}
\end{equation}

\subsection{Geometric Mean}
The loss function aiming at geometric mean, \textit{$L_{G}$}, is represented as (\ref{eq:gmean_appendix}) with the confusion matrix (\ref{eq:confusionmatrix}).
The first term of (\ref{eq:chainrule}) for \textit{$L_{G}$} is as (\ref{eq:gmeanloss_appendix}).
The partial derivative of \textit{$L_{G}$} is as (\ref{eq:gmeander}).

\begin{equation}
  \resizebox{0.9\hsize}{!}{$L_{G} = 1 - \sqrt{\frac{(\sum_{i=1}^{n} y_{i} \cdot yh_{i})(n - \sum_{i=1}^{n} y_{i} - \sum_{i=1}^{n} yh_{i} + \sum_{i=1}^{n} y_{i} \cdot yh_{i})}{(\sum_{i=1}^{n} y_{i})(n - \sum_{i=1}^{n} y_{i})}}$}
\label{eq:gmean_appendix}
\end{equation}

\begin{equation}
\begin{split}
  & \frac{\partial L_{G}} {\partial yh_{i}} = \frac{-2}{\sqrt{(\sum_{i=1}^{n} y_{i})(n - \sum_{i=1}^{n} y_{i})}} \times \\ 
  & [\frac{(\sum_{i=1}^{n} y_{i} \cdot yh'_{i})(n - \sum_{i=1}^{n} y_{i} - \sum_{i=1}^{n} yh_{i} + \sum_{i=1}^{n} y_{i} \cdot yh_{i})}{\sqrt{(\sum_{i=1}^{n} y_{i} \cdot yh_{i})(n - \sum_{i=1}^{n} y_{i} - \sum_{i=1}^{n} yh_{i} + \sum_{i=1}^{n} y_{i} \cdot yh_{i})}} + \\
  & \frac{(-\sum_{i=1}^{n} yh'_{i} + \sum_{i=1}^{n} y_{i}\cdot yh'_{i})(\sum_{i=1}^{n} y_{i}\cdot yh_{i}) }{\sqrt{(\sum_{i=1}^{n} y_{i} \cdot yh_{i})(n - \sum_{i=1}^{n} y_{i} - \sum_{i=1}^{n} yh_{i} + \sum_{i=1}^{n} y_{i} \cdot yh_{i})}}] \\
\end{split}
\label{eq:gmeanloss_appendix}
\end{equation}

\begin{equation}
\begin{split}
  & \frac{\partial L_{G}} {\partial \textbf{W}} = \frac{-2}{\sqrt{(\sum_{i=1}^{n} y_{i})(n - \sum_{i=1}^{n} y_{i})}} \times \\ 
  & [\frac{(\sum_{i=1}^{n} y_{i} \cdot \frac{dyh_{i}} {d\textbf{W}})(n - \sum_{i=1}^{n} y_{i} - \sum_{i=1}^{n} yh_{i} + \sum_{i=1}^{n} y_{i} \cdot yh_{i})}{\sqrt{(\sum_{i=1}^{n} y_{i} \cdot yh_{i})(n - \sum_{i=1}^{n} y_{i} - \sum_{i=1}^{n} yh_{i} + \sum_{i=1}^{n} y_{i} \cdot yh_{i})}} + \\
  & \frac{(-\sum_{i=1}^{n} \frac{dyh_{i}} {d\textbf{W}} + \sum_{i=1}^{n} y_{i}\cdot \frac{dyh_{i}} {d\textbf{W}})(\sum_{i=1}^{n} y_{i}\cdot yh_{i}) }{\sqrt{(\sum_{i=1}^{n} y_{i} \cdot yh_{i})(n - \sum_{i=1}^{n} y_{i} - \sum_{i=1}^{n} yh_{i} + \sum_{i=1}^{n} y_{i} \cdot yh_{i})}}] \\
\end{split}
\label{eq:gmeander}
\end{equation}

\subsection{Balanced Accuracy}
The loss function aiming at balanced accuracy, \textit{$L_{B}$}, is represented as (\ref{eq:baccu_appendix}) with the confusion matrix (\ref{eq:confusionmatrix}).
The first term of (\ref{eq:chainrule}) for \textit{$L_{B}$} is as (\ref{eq:bacculoss_appendix}).
The partial derivative of \textit{$L_{B}$} is as (\ref{eq:baccuder}).

\begin{equation}
  \resizebox{0.9\hsize}{!}{$L_{B} = 1 - \frac{n(\sum_{i=1}^{n} y_{i} \cdot yh_{i}) + n\sum_{i=1}^{n} y_{i} - \sum_{i=1}^{n} y_{i} \cdot \sum_{i=1}^{n} yh_{i} - (\sum_{i=1}^{n} y_{i})^{2}}{2(\sum_{i=1}^{n} y_{i})(n - \sum_{i=1}^{n} y_{i})}$}
\label{eq:baccu_appendix}
\end{equation}

\begin{equation}
\begin{split}
  & \frac{\partial L_{B}} {\partial yh_{i}} = \frac{-n(\sum_{i=1}^{n} y_{i} \cdot yh'_{i}) + \sum_{i=1}^{n} y_{i} \cdot \sum_{i=1}^{n} yh'_{i}}{2(\sum_{i=1}^{n} y_{i})(n - \sum_{i=1}^{n} y_{i})}\\
\end{split}
\label{eq:bacculoss_appendix}
\end{equation}

\begin{equation}
  \frac{\partial L_{B}} {\partial \textbf{W}} = \frac{-n(\sum_{i=1}^{n} y_{i} \cdot \frac{dyh_{i}} {d\textbf{W}}) + \sum_{i=1}^{n} y_{i} \cdot \sum_{i=1}^{n} \frac{dyh_{i}} {d\textbf{W}}}{2(\sum_{i=1}^{n} y_{i})(n - \sum_{i=1}^{n} y_{i})}
\label{eq:baccuder}
\end{equation}
\section{Results with the Details}  \label{App_B}
Table~\ref{tab:102data_appendix} shows the description of 102 datasets.
Table~\ref{tab:slp102data_appendix} and Table~\ref{tab:mlp102data_appendix} show the winning number and probability with the 102 datasets.
Table~\ref{tab:slp4data_appendix} and Table~\ref{tab:mlp4data_appendix} show the scores in the 4 datasets.

\begin{table*}[ht]
\small
  \caption{Description of 102 Diverse Datasets. The number of samples, the number of features, and the imbalance ratios.}
  \begin{tabular}{cccc|cccc|cccc|cccc}
    \toprule
    Data & \#Samp. & \#Feat. & Imb. & Data & \#Samp. & \#Feat. & Imb. & Data & \#Samp. & \#Feat. & Imb. & Data & \#Samp. & \#Feat. & Imb.\\
    \midrule
    1	&250	&12	&0.64:0.36	& 27	&504	&19	&0.91:0.09	& 53	&1000	&19	&0.70:0.30	& 78	&2201	&2	&0.68:0.32	\\
    2	&250	&9	&0.62:0.38	& 28	&522	&20	&0.80:0.20	& 54	&1000	&20	&0.74:0.26	& 79	&2310	&17	&0.86:0.14	\\
    3	&306	&3	&0.74:0.26	& 29	&531	&101&0.90:0.10	& 55	&1043	&37	&0.88:0.12	& 80	&2407	&299&0.82:0.18	\\
    4	&310	&6	&0.68:0.32	& 30	&540	&20	&0.91:0.09	& 56	&1055	&32	&0.66:0.34	& 81	&2417	&115&0.74:0.26	\\
    5	&320	&6	&0.67:0.33	& 31	&559	&4	&0.86:0.14	& 57	&1066	&7	&0.83:0.17	& 82	&2534	&71	&0.94:0.06	\\
    6	&327	&37	&0.87:0.13	& 32	&562	&21	&0.84:0.16	& 58	&1074	&16	&0.68:0.32	& 83	&3103	&12	&0.93:0.07	\\
    7	&328	&32	&0.69:0.31	& 33	&569	&30	&0.63:0.37	& 59	&1077	&37	&0.88:0.12	& 84	&3103	&12	&0.92:0.08	\\
    8	&335	&3	&0.85:0.15	& 34	&583	&10	&0.71:0.29	& 60	&1109	&21	&0.93:0.07	& 85	&4052	&5	&0.76:0.24	\\
    9	&336	&14	&0.76:0.24	& 35	&593	&77	&0.68:0.32	& 61	&1156	&5	&0.78:0.22	& 86	&4474	&11	&0.75:0.25	\\
    10	&349	&31	&0.62:0.38	& 36	&600	&61	&0.83:0.17	& 62	&1320	&17	&0.91:0.09	& 87	&4521	&14	&0.88:0.12	\\
    11	&351	&33	&0.64:0.36	& 37	&609	&7	&0.63:0.37	& 63	&1324	&10	&0.78:0.22	& 88	&4601	&7	&0.61:0.39	\\
    12	&358	&31	&0.69:0.31	& 38	&641	&19	&0.68:0.32	& 64	&1458	&37	&0.88:0.12	& 89	&4859	&120&0.83:0.17	\\
    13	&363	&8	&0.77:0.23	& 39	&645	&168&0.94:0.06	& 65	&1563	&37	&0.90:0.10	& 90	&5000	&40	&0.66:0.34	\\
    14	&365	&5	&0.92:0.08	& 40	&661	&37	&0.92:0.08	& 66	&1728	&6	&0.70:0.30	& 91	&5000	&19	&0.86:0.14	\\
    15	&381	&38	&0.85:0.15	& 41	&683	&9	&0.65:0.35	& 67	&1941	&31	&0.65:0.35	& 92	&5404	&5	&0.71:0.29	\\
    16	&392	&8	&0.62:0.38	& 42	&705	&37	&0.91:0.09	& 68	&2000	&6	&0.90:0.10	& 93	&5473	&10	&0.90:0.10	\\
    17	&400	&5	&0.78:0.22	& 43	&748	&4	&0.76:0.24	& 69	&2000	&76	&0.90:0.10	& 94	&5620	&48	&0.90:0.10	\\
    18	&403	&35	&0.92:0.08	& 44	&768	&8	&0.65:0.35	& 70	&2000	&216&0.90:0.10	& 95	&6598	&169&0.85:0.15	\\
    19	&450	&3	&0.88:0.12	& 45	&797	&4	&0.81:0.19	& 71	&2000	&47	&0.90:0.10	& 96	&6598	&168&0.85:0.15	\\
    20	&458	&38	&0.91:0.09	& 46	&812	&6	&0.77:0.23	& 72	&2000	&64	&0.90:0.10	& 97	&7032	&36	&0.89:0.11	\\
    21	&462	&9	&0.65:0.35	& 47	&841	&70	&0.62:0.38	& 73	&2000	&239&0.90:0.10	& 98	&7970	&39	&0.93:0.07	\\
    22	&470	&13	&0.85:0.15	& 48	&846	&18	&0.74:0.26	& 74	&2000	&139&0.72:0.28	& 99	&8192	&12	&0.70:0.30	\\
    23	&475	&3	&0.87:0.13	& 49	&958	&9	&0.65:0.35	& 75	&2000	&140&0.87:0.13	& 100	&8192	&19	&0.70:0.30	\\
    24	&475	&3	&0.87:0.13	& 50	&959	&40	&0.64:0.36	& 76	&2001	&2	&0.76:0.24	& 101	&8192	&32	&0.69:0.31	\\
    25	&500	&25	&0.61:0.39	& 51	&973	&9	&0.67:0.33	& 77	&2109	&20	&0.85:0.15	& 102	&9961	&14	&0.84:0.16	\\
    26	&500	&22	&0.84:0.16	& 52	&990	&13	&0.91:0.09	&       &       &   &           &       &       &   &           \\
  \bottomrule
\end{tabular}
\label{tab:102data_appendix}
\end{table*}

\begin{table*}[ht]
  \caption{Results in SLP for the 102 datasets.}
  \begin{tabular}{cccccccc}
    \toprule
      & \multicolumn{3}{c}{Winning Numbers} & & \multicolumn{2}{c}{Winning Probability} &  \\
    \midrule
     Eval. & MSE & BCE & OURS & \multicolumn{2}{c}{Win/Draw/Lose (r:0.01)} & \multicolumn{2}{c}{Win/ Draw/Lose (r:0.05)}\\
     Metrics & lr:1e-2 & lr:1e-1 & L$_{Any}$(lr) & OURS vs. MSE & OURS vs. BCE & OURS vs. MSE & OURS vs. BCE \\
    \midrule
     Acc     &  13 &  20 & \textbf{69} | L$_{A}$(5e-3)     & 0.005/0.995/0.000 & 0.000/1.000/0.000 & 0.000/1.000/0.000 & 0.000/1.000/0.000 \\
     F-1     &  2 &  10 & \textbf{90} | L$_{F_{1}}$(1e-2)  & 1.000/0.000/0.000 & 1.000/0.000/0.000 & 0.999/0.001/0.000 & 0.759/0.241/0.000 \\
     G-Mean  &  2 &  5 & \textbf{95} | L$_{G}$(5e-3)       & 1.000/0.000/0.000 & 1.000/0.000/0.000 & 1.000/0.000/0.000 & 0.932/0.068/0.000 \\
     B-Acc   &  4 &  9 & \textbf{89} | L$_{B}$(5e-3)       & 1.000/0.000/0.000 & 1.000/0.000/0.000 & 0.931/0.069/0.000 & 0.139/0.861/0.000 \\
    \bottomrule
    \multicolumn{4}{l}{Epochs: 1,000}
  \end{tabular}
  \label{tab:slp102data_appendix}
\end{table*}

\begin{table*}[ht]
  \caption{Results in MLP for the 102 datasets.}
  \begin{tabular}{cccccccc}
    \toprule
      & \multicolumn{3}{c}{Winning Numbers} & & \multicolumn{2}{c}{Winning Probability} &  \\
    \midrule
     Eval. & MSE & BCE & OURS & \multicolumn{2}{c}{Win/Draw/Lose (r:0.01)} & \multicolumn{2}{c}{Win/ Draw/Lose (r:0.05)}\\
     Metrics & lr:5e-3 & lr:3e-3 & L$_{Any}$(lr) & OURS vs. MSE & OURS vs. BCE & OURS vs. MSE & OURS vs. BCE \\
    \midrule
     Acc     &  16 &  20 & \textbf{66} | L$_{A}$(5e-3)      & 0.000/1.000/0.000 & 0.000/1.000/0.000 & 0.000/1.000/0.000 & 0.000/1.000/0.000 \\
     F-1     &  5 &  3 & \textbf{94} | L$_{F_{1}}$(1e-3)    & 1.000/0.000/0.000 & 1.000/0.000/0.000 & 0.844/0.156/0.000 & 0.501/0.499/0.000 \\
     G-Mean  &  4 &  8 & \textbf{90} | L$_{G}$(1e-2)        & 1.000/0.000/0.000 & 1.000/0.000/0.000 & 0.696/0.304/0.000 & 0.689/0.311/0.000 \\
     B-Acc   &  4 &  4 & \textbf{94} | L$_{B}$(5e-3)        & 1.000/0.000/0.000 & 1.000/0.000/0.000 & 0.181/0.819/0.000 & 0.095/0.905/0.000 \\
    \bottomrule
    \multicolumn{8}{l}{1 Hidden Layer (2 nodes) / Activation: Sigmoid / Epochs: 100 / Batch Size: train data size $\times$ 5e-2 (5e-1 for L$_{G}$ \& L$_{B}$)}
  \end{tabular}
  \label{tab:mlp102data_appendix}
\end{table*}

\begin{table*}[ht]
\small
  \caption{Results in SLP for the 4 datasets.}
  \begin{tabular}{ccccccccccccc}
    \toprule
    Data & Eval. & MSE & BCE & L$_{A}$ & L$_{F_{1}}$ & L$_{F_{.5}}$ & L$_{F_{2}}$ & L$_{G}$ & L$_{B}$ & S$_{A}$ & S$_{F}$ & S$_{B}$\\
    \midrule
    \#1 & Metrics & lr:1e-2 & lr:2e-1 & lr:1e-2 & lr:1e-2 & lr:1e-2 & lr:1e-2 & lr:1e-2 & lr:1e-2 & lr:1e-2 & lr:5e-3 & lr:5e-3\\
    \midrule
              & Acc   &	0.898000 &	0.921500 &	0.922000 &	\textcolor{lightgray}{0.913900} &	\textcolor{lightgray}{0.923500} &	\textcolor{lightgray}{0.872600} &	\textcolor{lightgray}{0.859400} &	\textcolor{lightgray}{0.853300} &0.922400 &\textcolor{lightgray}{0.913000} &\textcolor{lightgray}{0.853300}\\
       Ran.   & F-1   &	0.075428 &	0.546856 &	\textcolor{lightgray}{0.540208} &	0.632383 &	\textcolor{lightgray}{0.572610} &	\textcolor{lightgray}{0.589266} &	\textcolor{lightgray}{0.571078} &	\textcolor{lightgray}{0.563108} &\textcolor{lightgray}{0.533386} &0.634453 &\textcolor{lightgray}{0.562872}\\
      Gen.   & G-Mean&	0.194819 &	0.664798 &	\textcolor{lightgray}{0.654791} &	\textcolor{lightgray}{0.814538} &	\textcolor{lightgray}{0.690798} &	\textcolor{lightgray}{0.872564} &	0.874335 &	\textcolor{lightgray}{0.874540} &\textcolor{lightgray}{0.646546} &\textcolor{lightgray}{0.821146} &\textcolor{lightgray}{0.874122}\\
              & B-Acc &	0.519203 &	0.714945 &	\textcolor{lightgray}{0.708910} &	\textcolor{lightgray}{0.823336} &	\textcolor{lightgray}{0.732524} &	\textcolor{lightgray}{0.872757} &	\textcolor{lightgray}{0.874674} &	0.875064 &\textcolor{lightgray}{0.705226} &\textcolor{lightgray}{0.828751} &0.874640\\
    \midrule
    \#2 & Metrics & lr:5e-3 & lr:1e-2 & lr:1e-3 & lr:5e-3 & lr:5e-3 & lr:5e-4 & lr:1e-3 & lr:5e-3 & lr:5e-3 & lr:1e-3 & lr:5e-3\\
    \midrule
            & Acc   & 0.989800 & 0.990078 & 0.990430 & \textcolor{lightgray}{0.991572} & \textcolor{lightgray}{0.991210} & \textcolor{lightgray}{0.992128} & \textcolor{lightgray}{0.981580} & \textcolor{lightgray}{0.982196} &0.990832 &\textcolor{lightgray}{0.991016} &\textcolor{lightgray}{0.984491}\\
      Cre.  & F-1   & 0.880601 & 0.884281 & \textcolor{lightgray}{0.888906} & 0.903505 & \textcolor{lightgray}{0.898935} & \textcolor{lightgray}{0.913576} & \textcolor{lightgray}{0.827887} & \textcolor{lightgray}{0.832658} &\textcolor{lightgray}{0.894053} &0.896437 &\textcolor{lightgray}{0.851334}\\
    Card   & G-Mean& 0.888696 & 0.892165 & \textcolor{lightgray}{0.896594} & \textcolor{lightgray}{0.910165} & \textcolor{lightgray}{0.905946} & \textcolor{lightgray}{0.933853} & 0.956730 & \textcolor{lightgray}{0.956942} &\textcolor{lightgray}{0.901414} &\textcolor{lightgray}{0.903675} &\textcolor{lightgray}{0.956256}\\
    	  & B-Acc & 0.894874 & 0.897960 & \textcolor{lightgray}{0.901920} & \textcolor{lightgray}{0.914181} & \textcolor{lightgray}{0.910349} & \textcolor{lightgray}{0.935924} & \textcolor{lightgray}{0.957116} & 0.957340 &\textcolor{lightgray}{0.906273} &\textcolor{lightgray}{0.908308} &0.956908\\
    \midrule
    \#3 & Metrics & lr:5e-2 & lr:5e-2 & lr:5e-4 & lr:1e-3 & lr:1e-2 & lr:7e-3 & lr:3e-3 & lr:5e-3 & lr:1e-4 & lr:1e-4 & lr:1e-4\\
    \midrule
            & Acc	& 0.979615 & 0.984679 & 0.987244 & \textcolor{lightgray}{0.984679} & \textcolor{lightgray}{0.984679} & \textcolor{lightgray}{0.989744} & \textcolor{lightgray}{0.982179} & \textcolor{lightgray}{0.982179} &0.979808 &\textcolor{lightgray}{0.982372} &\textcolor{lightgray}{0.977244}\\
     Bre.   & F-1	& 0.863810 & 0.901270 & \textcolor{lightgray}{0.912381} & 0.901270 & \textcolor{lightgray}{0.892381} & \textcolor{lightgray}{0.932381} & \textcolor{lightgray}{0.901270} & \textcolor{lightgray}{0.906032} &\textcolor{lightgray}{0.908045} &0.922330 &\textcolor{lightgray}{0.893759}\\
     Can.   & G-Mean& 0.875467 & 0.919174 & \textcolor{lightgray}{0.920613} & \textcolor{lightgray}{0.919174} & \textcolor{lightgray}{0.902262} & \textcolor{lightgray}{0.950962} & 0.946705 & \textcolor{lightgray}{0.961198} &\textcolor{lightgray}{0.964899} &\textcolor{lightgray}{0.966298} &\textcolor{lightgray}{0.963500}\\
            & B-Acc	& 0.887500 & 0.927738 & \textcolor{lightgray}{0.929167} & \textcolor{lightgray}{0.927738} & \textcolor{lightgray}{0.912500} & \textcolor{lightgray}{0.956944} & \textcolor{lightgray}{0.952738} & 0.963849 &\textcolor{lightgray}{0.966627} &\textcolor{lightgray}{0.968016} &0.965238\\
    \midrule
    \#4 & Metrics & lr:5e-2 & lr:1e-1 & lr:1e-2 & lr:1e-2 & lr:1e-2 & lr:1e-2 & lr:1e-2 & lr:1e-2 & lr:1e-2 & lr:1e-2 & lr:1e-3\\
    \midrule
            & Acc	& 0.939810 & 0.959400 & 0.960300 & \textcolor{lightgray}{0.959950} & \textcolor{lightgray}{0.959350} & \textcolor{lightgray}{0.924600} & \textcolor{lightgray}{0.878080} & \textcolor{lightgray}{0.875210} &0.897990 &\textcolor{lightgray}{0.942200} &\textcolor{lightgray}{0.871320}\\
      Dia.   & F-1	& 0.453943 & 0.710881 & \textcolor{lightgray}{0.708927} & 0.732548 & \textcolor{lightgray}{0.690117} & \textcolor{lightgray}{0.647293} & \textcolor{lightgray}{0.555013} & \textcolor{lightgray}{0.550022} &\textcolor{lightgray}{0.543939} &0.670434 &\textcolor{lightgray}{0.541369}\\
     Pred   & G-Mean& 0.542504 & 0.764060 & \textcolor{lightgray}{0.752976} & \textcolor{lightgray}{0.798967} & \textcolor{lightgray}{0.729499} & \textcolor{lightgray}{0.872160} & 0.885350 & \textcolor{lightgray}{0.884975} &\textcolor{lightgray}{0.809091} &\textcolor{lightgray}{0.813986} &\textcolor{lightgray}{0.879566}\\
            & B-Acc	& 0.647168 & 0.790684 & \textcolor{lightgray}{0.782798} & \textcolor{lightgray}{0.817291} & \textcolor{lightgray}{0.765898} & \textcolor{lightgray}{0.874277} & \textcolor{lightgray}{0.885407} & 0.885066 &\textcolor{lightgray}{0.815235} &\textcolor{lightgray}{0.826107} &0.879686\\
  \bottomrule
  \multicolumn{10}{l}{Epochs: 1,000}\\
  \multicolumn{13}{l}{Settings for SOL = S$_{A}$(cosine/mu:0.5/delta=0.1) / S$_{F}$(cosine/mu:0.5/delta=0.1) / S$_{B}$(cosine/mu:0.5/delta=0.1)  / Epoch: 100} \\
\end{tabular}
\label{tab:slp4data_appendix}
\end{table*}

\begin{table*}[ht]
\small
  \caption{Results in MLP for the 4 datasets.}
  \begin{tabular}{ccccccccccccc}
    \toprule
    Data & Eval. & MSE & BCE & L$_{A}$ & L$_{F_{1}}$ & L$_{F_{.5}}$ & L$_{F_{2}}$ & L$_{G}$ & L$_{B}$ & S$_{A}$ & S$_{F}$ & S$_{B}$\\
    \midrule
    \#1 & Metrics & lr:1e-3 & lr:5e-3 & lr:4e-3 & lr:1e-3 & lr:5e-3 & lr:5e-3 & lr:5e-3 & lr:5e-3 & lr:1e-3 & lr:1e-3 & lr:2e-2\\
    \midrule
             & Acc   & 0.920300 & 0.923900 & 0.924500 & \textcolor{lightgray}{0.916500} & \textcolor{lightgray}{0.924900} & \textcolor{lightgray}{0.877700} & \textcolor{lightgray}{0.834700} & \textcolor{lightgray}{0.876400} & 0.923100 & \textcolor{lightgray}{0.914600} & \textcolor{lightgray}{0.827200}\\
        Ran. & F-1   & 0.527334 & 0.590095 & \textcolor{lightgray}{0.578506} & 0.636631 & \textcolor{lightgray}{0.584536} & \textcolor{lightgray}{0.594422} & \textcolor{lightgray}{0.540051} & \textcolor{lightgray}{0.591280} & \textcolor{lightgray}{0.557660} & 0.636488 & \textcolor{lightgray}{0.531008}\\
       Gen.  & G-Mean& 0.650637 & 0.712684 & \textcolor{lightgray}{0.694489} & \textcolor{lightgray}{0.811241} & \textcolor{lightgray}{0.703328} & \textcolor{lightgray}{0.867475} & 0.866778 & \textcolor{lightgray}{0.859828} & \textcolor{lightgray}{0.672888} & \textcolor{lightgray}{0.818277} & \textcolor{lightgray}{0.856774}\\
             & B-Acc & 0.711984 & 0.747529 & \textcolor{lightgray}{0.735166} & \textcolor{lightgray}{0.820982} & \textcolor{lightgray}{0.742470} & \textcolor{lightgray}{0.868011} & \textcolor{lightgray}{0.868504} & 0.861911 & \textcolor{lightgray}{0.720869} & \textcolor{lightgray}{0.826697} & 0.857939\\
    \midrule
    \#2 & Metrics & lr:5e-4 & lr:5e-3 & lr:3e-3 & lr:3e-3 & lr:3e-3 & lr:3e-3 & lr:3e-3 & lr:5e-3 & lr:5e-3 & lr:1e-2 & lr:5e-3\\
    \midrule
             & Acc   & 0.990567 & 0.993532 & 0.993997 & \textcolor{lightgray}{0.995066} & \textcolor{lightgray}{0.992155} & \textcolor{lightgray}{0.994369} & \textcolor{lightgray}{0.977419} & \textcolor{lightgray}{0.990741} & 0.992095 & \textcolor{lightgray}{0.994868} & \textcolor{lightgray}{0.981027}\\
     Cre.    & F-1   & 0.888571 & 0.928710 & \textcolor{lightgray}{0.933364} & 0.946277 & \textcolor{lightgray}{0.910775} & \textcolor{lightgray}{0.940188} & \textcolor{lightgray}{0.804384} & \textcolor{lightgray}{0.904573} & \textcolor{lightgray}{0.910032} & 0.944155 & \textcolor{lightgray}{0.821155}\\
     Card    & G-Mean& 0.897839 & 0.940434 & \textcolor{lightgray}{0.941921} & \textcolor{lightgray}{0.955223} & \textcolor{lightgray}{0.916897} & \textcolor{lightgray}{0.962974} & 0.960083 & \textcolor{lightgray}{0.956059} & \textcolor{lightgray}{0.916242} & \textcolor{lightgray}{0.954546} & \textcolor{lightgray}{0.948832}\\
             & B-Acc & 0.904099 & 0.942173 & \textcolor{lightgray}{0.943752} & \textcolor{lightgray}{0.956211} & \textcolor{lightgray}{0.920334} & \textcolor{lightgray}{0.963597} & \textcolor{lightgray}{0.960347} & 0.956882 & \textcolor{lightgray}{0.919734} & \textcolor{lightgray}{0.955573} & 0.949542\\
    \midrule
    \#3 & Metrics & lr:1e-3 & lr:5e-3 & lr:5e-3 & lr:1e-2 & lr:1e-2 & lr:3e-3 & lr:1e-2 & lr:5e-3 & lr:1e-3 & lr:1e-2 & lr:5e-3\\
    \midrule
             & Acc   & 0.982244 & 0.994872 & 0.994936 & \textcolor{lightgray}{1.000000} & \textcolor{lightgray}{0.994936} & \textcolor{lightgray}{1.000000} & \textcolor{lightgray}{0.994872} & \textcolor{lightgray}{0.969872} & 0.984679 & \textcolor{lightgray}{0.992372} & \textcolor{lightgray}{0.982179}\\
     Bre.    & F-1   & 0.882857 & 0.965714 & \textcolor{lightgray}{0.974603} & 1.000000 & \textcolor{lightgray}{0.965714} & \textcolor{lightgray}{1.000000} & \textcolor{lightgray}{0.968889} & \textcolor{lightgray}{0.910551} & \textcolor{lightgray}{0.890317} & 0.954603 & \textcolor{lightgray}{0.906984}\\
     Can.    & G-Mean& 0.891359 & 0.968252 & \textcolor{lightgray}{0.985164} & \textcolor{lightgray}{1.000000} & \textcolor{lightgray}{0.968252} & \textcolor{lightgray}{1.000000} & 0.980211 & \textcolor{lightgray}{0.982366} & \textcolor{lightgray}{0.911151} & \textcolor{lightgray}{0.966813} & \textcolor{lightgray}{0.973177}\\
             & B-Acc & 0.900000 & 0.970833 & \textcolor{lightgray}{0.986071} & \textcolor{lightgray}{1.000000} & \textcolor{lightgray}{0.970833} & \textcolor{lightgray}{1.000000} & \textcolor{lightgray}{0.981905} & 0.983294 & \textcolor{lightgray}{0.923571} & \textcolor{lightgray}{0.969405} & 0.974921\\
    \midrule
    \#4 & Metrics & lr:1e-3 & lr:5e-3 & lr:5e-3 & lr:3e-3 & lr:5e-3 & lr:3e-3 & lr:1e-2 & lr:7e-2 & lr:1e-2 & lr:2e-2 & lr:3e-2\\
    \midrule
             & Acc   & 0.955990 & 0.960370 & 0.961340 & \textcolor{lightgray}{0.960310} & \textcolor{lightgray}{0.960760} & \textcolor{lightgray}{0.938710} & \textcolor{lightgray}{0.862870} & \textcolor{lightgray}{0.926500} & 0.961200 & \textcolor{lightgray}{0.960130} & \textcolor{lightgray}{0.862430}\\
       Dia.   & F-1   & 0.654638 & 0.716911 & \textcolor{lightgray}{0.724130} & 0.736366 & \textcolor{lightgray}{0.703492} & \textcolor{lightgray}{0.691992} & \textcolor{lightgray}{0.536563} & \textcolor{lightgray}{0.662707} & \textcolor{lightgray}{0.724056} & 0.735649 & \textcolor{lightgray}{0.519180}\\
     Pred    & G-Mean& 0.714728 & 0.766348 & \textcolor{lightgray}{0.770847} & \textcolor{lightgray}{0.803085} & \textcolor{lightgray}{0.739861} & \textcolor{lightgray}{0.877184} & 0.887109 & \textcolor{lightgray}{0.874578} & \textcolor{lightgray}{0.771942} & \textcolor{lightgray}{0.803348} & \textcolor{lightgray}{0.864478}\\
             & B-Acc & 0.765876 & 0.792601 & \textcolor{lightgray}{0.796173} & \textcolor{lightgray}{0.820582} & \textcolor{lightgray}{0.773552} & \textcolor{lightgray}{0.880067} & \textcolor{lightgray}{0.888034} & 0.877450 & \textcolor{lightgray}{0.796950} & \textcolor{lightgray}{0.820751} & 0.865170\\
  \bottomrule
  \multicolumn{13}{l}{1 Hidden Layer (2 nodes) / Activation: Sigmoid / Epochs: 100 / Batch Size: train data size $\times$ 5e-2 (5e-1 for L$_{G}$ \& L$_{B}$)} \\
  \multicolumn{13}{l}{Settings for SOL = S$_{A}$(cosine/mu:0.5/delta=0.1) / S$_{F}$(cosine/mu:0.5/delta=0.1) / S$_{B}$(cosine/mu:0.5/delta=0.1) } \\
\end{tabular}
\label{tab:mlp4data_appendix}
\end{table*}

\end{document}